\RequirePackage{letltxmacro}
\documentclass[pdflatex,sn-mathphys]{sn-jnl}

\jyear{2021}%

\theoremstyle{thmstyleone}%
\usepackage[T5]{fontenc}
\usepackage[utf8]{inputenc}
\usepackage{booktabs}
\usepackage{float}
\usepackage{adjustbox}
\usepackage{caption}
\usepackage{subcaption}
\usepackage{graphicx}
\usepackage{amsmath}
\usepackage{algorithm}
\usepackage{algpseudocode}
\usepackage{threeparttable}
\usepackage{vcell}
\usepackage{xcolor}
\usepackage[dvipsnames]{xcolor}
\usepackage[vietnamese,english]{babel}

\usepackage{longtable}

\usepackage{multirow}
\usepackage{caption}
\usepackage{makecell}
\usepackage{graphicx}%
\usepackage{multirow}%

\usepackage{array}
\usepackage{adjustbox}

\usepackage{amssymb}
\usepackage{pifont}

\theoremstyle{thmstyletwo}%

\theoremstyle{thmstylethree}%

\usepackage{soul}
\renewcommand{\hl}[1]{#1} 

\definecolor{lavender}{RGB}{240,230,250}
\definecolor{lightblue}{RGB}{173, 216, 230}

\sethlcolor{yellow}

\begin{document}


\title[Tin et al.]{ViCLSR: A Supervised Contrastive Learning Framework with Natural Language Inference for Natural Language Understanding Tasks}



\author[1,2]{\fnm{Tin} \sur{Van Huynh}}\email{tinhv@uit.edu.vn}

\author*[1,2]{\fnm{Kiet} \sur{Van Nguyen}}\email{kietnv@uit.edu.vn}

\author[1,2]{\fnm{Ngan} \sur{Luu-Thuy Nguyen}}\email{ngannlt@uit.edu.vn}

\affil[1]{\orgname{University of Information Technology, Ho Chi Minh City, Vietnam}}
\affil[2]{\orgname{Vietnam National University, Ho Chi Minh City, Vietnam}}


\abstract{
High-quality text representations are crucial for natural language understanding (NLU), but low-resource languages like Vietnamese face challenges due to limited annotated data. While pre-trained models like PhoBERT and CafeBERT perform well, their effectiveness is constrained by data scarcity. Contrastive learning (CL) has recently emerged as a promising approach for improving sentence representations, enabling models to effectively distinguish between semantically similar and dissimilar sentences. We propose ViCLSR (Vietnamese Contrastive Learning for Sentence Representations), a novel supervised contrastive learning framework specifically designed to optimize sentence embeddings for Vietnamese, leveraging existing natural language inference (NLI) datasets. Additionally, we propose a process to adapt existing Vietnamese datasets for supervised learning, ensuring compatibility with CL methods. Our experiments demonstrate that ViCLSR significantly outperforms the powerful monolingual pre-trained model PhoBERT on five benchmark NLU datasets such as ViNLI (+6.97\% F1), ViWikiFC (+4.97\% F1), ViFactCheck (+9.02\% F1), UIT-ViCTSD (+5.36\% F1), and ViMMRC2.0 (+4.33\% Accuracy). ViCLSR shows that supervised contrastive learning can effectively address resource limitations in Vietnamese NLU tasks and improve sentence representation learning for low-resource languages. Furthermore, we conduct an in-depth analysis of the experimental results to uncover the factors contributing to the superior performance of contrastive learning models. ViCLSR is released for research purposes\footnote{We will provide an access link to it as soon as the article is accepted.}\hspace{0.3em} in advancing natural language processing tasks.
}




\maketitle

\section{Introduction}
\label{sect:introduction}

Natural Language Understanding (NLU) has become a critical component in a wide range of natural language processing (NLP) applications, as it enables machines to interpret and reason about human language in a way that aligns with human cognitive processes. Recently, contrastive learning has proven effective for improving the quality of sentence embeddings, which are crucial for NLU tasks \cite{gao2021simcse}. By contrasting positive and negative sentence pairs, contrastive learning optimizes the ability of model to understand subtle semantic relationships between sentences, which is crucial for tasks such as natural language inference and fact checking \cite{van2022vinli, hoa2024vifactcheck}. This technique has been proven to significantly improve the performance of models in tasks that require deep semantic reasoning and contextual understanding \cite{wang2022english, chuang2022diffcse}. Moreover, the adaptability of contrastive learning allows it to be effectively utilized in various NLU tasks, further enhancing its versatility in addressing diverse challenges.

Text representation is fundamental to NLU tasks, transforming raw text into dense vector embeddings that capture semantic meaning. These embeddings serve as the foundation for a wide array of NLU tasks, including natural language inference \cite{bowman2015large}, semantic similarity detection \cite{cer2017semeval}, fact checking \cite{thorne2018fever}, machine reading comprehension (MRC) \cite{lai2017race}, and text classification \cite{howard2018universal}. While substantial progress has been made in high-resource languages like English, NLU for low-resource languages such as Vietnamese remains a significant challenge. This is due to the unique linguistic properties of Vietnamese, such as its complex syntax, morphology, and semantics, which are often inadequately captured by existing models.

In recent years, transformer-based pre-trained language models have revolutionized NLU, with multilingual models like mBERT \cite{devlin2019bert} and XLM-R \cite{conneau-etal-2020-unsupervised} showcasing exceptional performance across diverse languages. However, these models are trained on textual data spanning hundreds of languages, leading to diluted focus on Vietnamese-specific characteristics. Studies \cite{van2022vinli, le2024viwikifc, hoa2024vifactcheck} have shown that such multilingual models often struggle to capture subtle linguistic nuances, particularly in complex tasks like NLI and fact checking.

Besides, monolingual models like PhoBERT \cite{nguyen2020phobert}, tailored specifically for Vietnamese, leverage the RoBERTa \cite{liuroberta} architecture and a large corpus of Vietnamese text. These models have demonstrated notable improvements over multilingual counterparts in specific tasks, as shown in research by Nguyen et al. (2020) \cite{nguyen2020phobert}, and Do et al. (2024) \cite{do2024vlue}. However, their ability to represent complex semantic relationships remains limited. This limitation is particularly evident in tasks requiring profound contextual understanding, such as natural language inference \cite{van2022vinli, van2022error}, machine reading comprehension \cite{luu2023multiple}, and fact checking \cite{le2024viwikifc}, where PhoBERT still falls short compared to cutting-edge pre-trained models. Addressing these challenges requires more advanced approaches that can fully leverage the unique linguistic characteristics of the Vietnamese language, especially in resource-constrained settings.

In this study, we propose a novel supervised contrastive learning framework to enhance NLU for Vietnamese. By integrating contrastive learning with the XLM-R architecture, we aim to improve the ability of the model to capture complex semantic relationships. We utilize existing Vietnamese NLI datasets and adapt them into contrastive-like datasets including positive sentence pairs (\( x \), \( x^{+} \)) and negative sentence pairs (\( x \), \( x^{-} \)) to be suitable for training contrastive learning models, allowing the model to optimize embeddings for improved semantic reasoning. Our proposed model demonstrates significant improvements over both multilingual and monolingual pre-trained models, achieving state-of-the-art results across a variety of Vietnamese NLU tasks, particularly those requiring profound contextual understanding, such as NLI, fact checking, machine reading comprehension, and constructive speech detection.

The four primary contributions of our research in this paper are summarized as follows.

\begin{itemize}
    \item \textbf{ViCLSR Framework}: We propose ViCLSR, a novel supervised contrastive learning framework designed to improve Vietnamese sentence embeddings. By integrating contrastive learning with the XLM-R architecture, ViCLSR leverages existing NLI datasets (ViNLI and XNLI) and introduces a data adaptation process to create positive and negative sentence pairs, enhancing semantic representation for NLU tasks.
    
    \item \textbf{Empirical Evaluation}: We conduct a comprehensive evaluation of the effectiveness of ViCLSR through extensive experiments on five benchmark datasets: ViNLI, ViWikiFC, ViFactCheck, UIT-ViCTSD, and ViMMRC2.0—covering tasks like natural language inference, fact checking, constructive speech detection, and machine reading comprehension. The experimental results demonstrate its superiority over existing multilingual (mBERT, XLM-R) and monolingual (PhoBERT) pre-trained models, highlighting its effectiveness in Vietnamese NLU.
    
    \item \textbf{In-Depth Analysis}: We provide a detailed analysis of the factors including hyperparameter studies (e.g., temperature $\tau$), ablation tests on auxiliary MLM objectives, pooling strategies, semantic representation capabilities, sentence embedding distributions, and attention mechanisms that contribute to the success of ViCLSR, enhancing the understanding of how contrastive learning can improve semantic representations for low-resource languages like Vietnamese.


    \item \textbf{Public Release}: We publicly release our pre-trained model (ViCLSR) to support reproducibility and future research. This open-access resource is especially valuable for low-resource language communities, enabling researchers to build upon or compare with ViCLSR in advancing NLU applications.

\end{itemize}

The structure of our paper is organized as follows. Section \ref{sect:introduction} introduces text representation and highlights the challenges they pose for low-resource languages like Vietnamese. Section \ref{sect:related_works} discusses related works, including text representation, contrastive learning, training data resource for contrastive learning. Section \ref{sect:ViCSE} presents our ViCLSR approach, covering data preparation, training with contrastive learning, and fine-tuning for downstream tasks. Section \ref{sect:experiment} presents the experiments and results. Section \ref{sect:result_analysis} provides an analysis and discussion of the findings. Finally, Section \ref{sect:conclusion} concludes the paper by summarizing the key research contributions and proposing directions for future work.

\begin{figure*}[h!]
    \centering
    \includegraphics[width=1\linewidth]{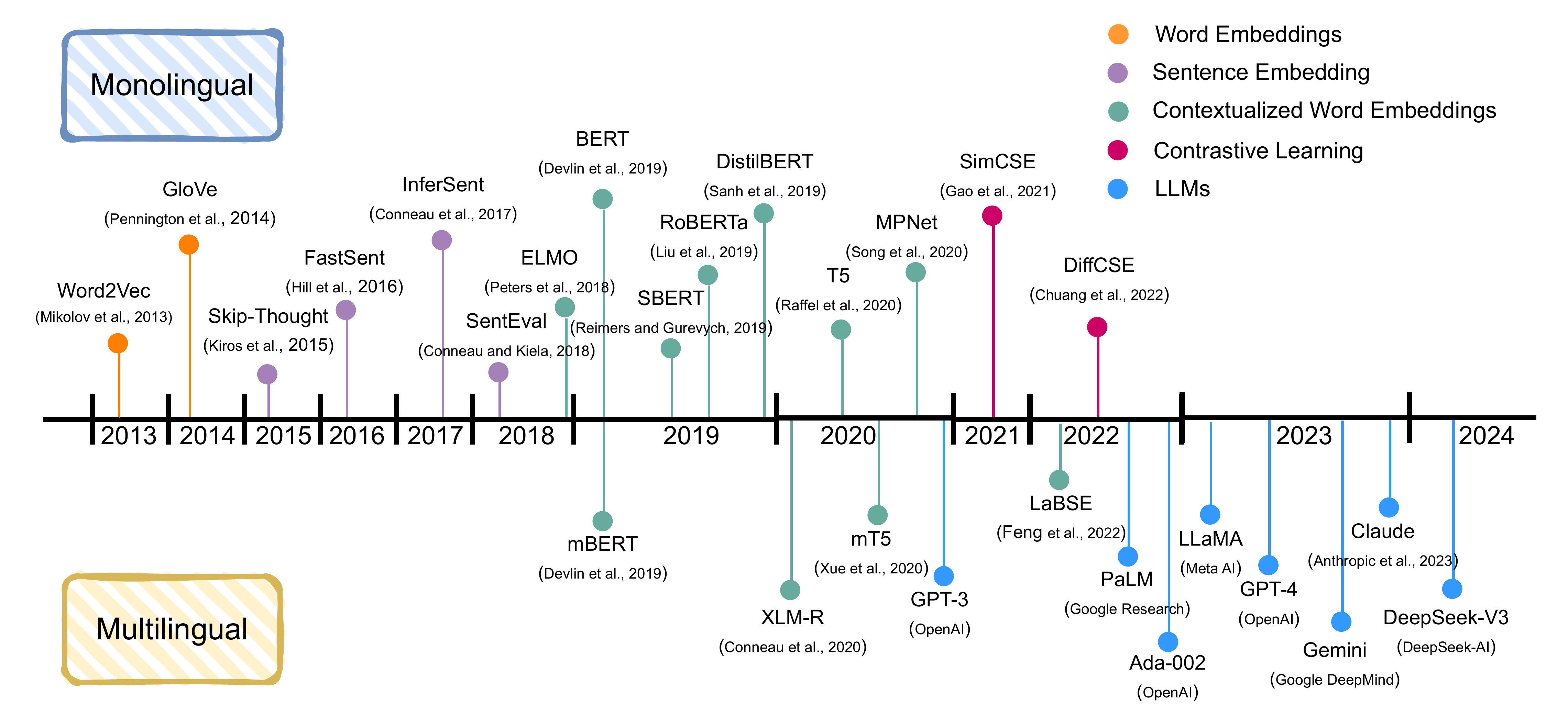}
    \caption{Overview of Typical Text Representation Models from 2013 to 2024.}
    \label{fig:Text-Representation-Overview}
\end{figure*}

\section{Related Works}
\label{sect:related_works}

We provide a review of related work, outlining the evolution of text representation from basic methods to advanced transformer-based models, and discuss how contrastive learning improves sentence representations, especially for NLU tasks requiring semantic understanding.

\subsection{Text Representation}


Text representation plays a crucial role in natural language processing, as it transforms textual data into numerical representations that capture semantic meaning. As illustrated in Figure \ref{fig:Text-Representation-Overview}, the evolution of text representation methods has progressed from early statistical approaches to modern transformer-based architectures. Initial techniques such as bag-of-words (BoW) and TF-IDF treated words as independent units, lacking contextual awareness and semantic relationships. The introduction of word embeddings, such as Word2Vec \cite{mikolov2013efficient} and GloVe \cite{pennington2014glove}, marked a significant shift by encoding words into continuous vector spaces based on co-occurrence patterns. These methods improved word-level representations but remained context-independent, limiting their ability to capture sentence-level semantics effectively.

To overcome these limitations, researchers developed Sentence Embeddings, where models like Skip-Thought \cite{kiros2015skip}, InferSent \cite{conneau2017supervised}, and FastSent \cite{hill2016learning} focused on generating sentence-level representations. These models improved upon word embeddings by considering larger linguistic contexts but were often constrained by their reliance on fixed-length vector representations and lack of task-specific fine-tuning.

A paradigm shift occurred with the advent of Contextualized Word Embeddings, pioneered by models such as ELMo \cite{peters-etal-2018-deep}, BERT \cite{devlin2019bert}, RoBERTa \cite{liuroberta}, SBERT \cite{reimers-2019-sentence-bert}, DistilBERT \cite{sanh2019distilbert}, and T5 \cite{raffel2020exploring}. These transformer-based architectures introduced self-attention mechanisms, enabling dynamic representations that capture word dependencies across different sentence positions. Unlike traditional embeddings, these models consider both left and right contexts, significantly improving performance in NLU tasks such as machine translation, text classification, and natural language inference.

Recent advancements have explored contrastive learning as a method for improving sentence representations. Models like SimCSE \cite{gao2021simcse} and DiffCSE \cite{chuang2022diffcse}) have leveraged contrastive objectives to align semantically similar sentences while pushing apart dissimilar ones. These approaches have demonstrated superior performance in semantic similarity, sentence clustering, and retrieval-based NLU tasks compared to traditional methods.

The rapid evolution of Large Language Models (LLMs) has further transformed text representation. Beginning with GPT-3 \cite{brown2020language}, followed by GPT-4 \cite{achiam2023gpt}, LLaMA \cite{touvron2023llama} , and Gemini \cite{team2023gemini} in 2023 or DeepSeek-V3 \cite{liu2024deepseek} in 2024, these models have moved beyond static text representations to context-aware, generative models capable of handling multi-turn dialogues and diverse NLP applications. The integration of pre-trained transformer architectures has made it possible to fine-tune these models for specific domains, optimizing their performance across various natural language understanding and generation tasks.

In the context of multilingual NLU, XLM-R \cite{conneau-etal-2020-unsupervised} and mBERT \cite{devlin2019bert} have significantly advanced cross-lingual transfer learning by enabling shared representations across multiple languages. Among these, XLM-R stands out due to its robust training on a massive multilingual corpus, which allows it to capture both syntactic and semantic nuances across diverse linguistic structures. This makes XLM-R particularly effective for low-resource languages like Vietnamese, as it benefits from rich cross-lingual knowledge transfer while maintaining strong performance on monolingual tasks. In Vietnamese NLU, specialized models such as PhoBERT \cite{nguyen2020phobert} and CafeBERT \cite{do2024vlue} have further refined language representations. Recent efforts have specifically tailored these architectures to improve Vietnamese natural language inference (NLI). For instance, integrating contextualized language models with deep neural network architectures has been shown to optimize classification performance on NLI benchmarks \cite{nguyen2024transformer}. Furthermore, enhancing these pre-trained models with external contextual knowledge retrieval has proven effective in addressing complex passage-level reasoning capabilities \cite{nguyen2025lmck}. However, despite these task-specific adaptations, existing approaches still face challenges in fine-grained semantic understanding, particularly in natural language inference and sentence similarity evaluation, where effective sentence embeddings are crucial.

To address this, supervised contrastive learning has emerged as a promising approach for enhancing sentence representations by leveraging labeled data to better structure the semantic space. In this research, we propose a supervised contrastive learning framework built upon XLM-R \cite{conneau-etal-2020-unsupervised}, aiming to optimize sentence embeddings for Vietnamese and extend them to other low-resource languages. By leveraging the strong multilingual encoding capabilities of XLM-R and refining its representations with contrastive learning objectives, this approach aims to enhance sentence-level semantic alignment and task-specific generalization. This framework has the potential to bridge performance gaps between high-resource and low-resource languages, making sentence embeddings more robust, adaptable, and semantically enriched for diverse NLP applications.

\subsection{Contrastive Learning}
\label{subsection_Contrastive Learning}
Contrastive learning has emerged as a transformative approach in NLU by focusing on learning meaningful sentence representations through the comparison of similar and dissimilar data points. This paradigm shift started with the introduction of models like SimCSE \cite{gao2021simcse}, which utilized contrastive learning to enhance sentence embeddings by leveraging both unsupervised and supervised training approaches. In SimCSE, the unsupervised approach generates positive pairs by applying dropout noise, while the supervised method uses natural language inference datasets as labeled pairs, showing that contrastive learning can bring significant improvements over traditional models like BERT.  The overall goal of contrastive learning is to minimize the distance between semantically similar sentences while maximizing the distance between dissimilar ones in the vector space, which makes it highly effective for tasks like semantic similarity, paraphrase detection, and sentence clustering.

Based on the effectiveness of contrastive learning in capturing semantic relationships within text, this approach has been widely applied across various NLU tasks, resulting in remarkable improvements. It has led to significant advancements in areas such as text classification \cite{wang2021cline, wang2022incorporating}, sentence embeddings and phrase embeddings \cite{gao2021simcse, kim2021self, li2022uctopic}, information extraction \cite{das2022container, liu2022hiure}, machine translation \cite{pan2021contrastive, li2022improving}, machine reading comprehension \cite{you2021self, ji2022answer}, and summarization \cite{cao2021cliff, wu2020unsupervised}. Additionally, contrastive learning has been heavily explored in computer vision \cite{chen2020simple}, where it is used to detect similarities in visual data, highlighting its adaptability across both text and image processing domains.

Contrastive Learning has become a vital method for improving representation learning in low-resource languages, which often lack large annotated datasets. In the broader context of low-resource languages, various models have been developed using CL to enhance semantic representations for languages like Khmer and Pashto \cite{tan2022bitext}, Bengali \cite{maisha2024study}, and Mongolian \cite{liu2023approach}. Additionally, there has been a notable amount of research on contrastive learning in multilingual settings, which has made valuable strides in supporting semantic representations for low-resource languages \cite{hu2023language, huang2023cross}. For Vietnamese, the number of studies related to contrastive learning remains limited. Among the existing research, some focus on specific tasks, such as Vietnamese abstractive summarization \cite{dang2023contrastive}, but these studies lack generalizability and are often not easily applicable to a wide range of NLU tasks. Another example includes research on cross-lingual sentence embeddings between Chinese and Vietnamese \cite{huang2023cross}, which, while showing promising results, lacks focus on fully leveraging the unique linguistic features of Vietnamese. This highlights the need for more dedicated research on contrastive learning specifically tailored to Vietnamese, aiming to enhance its applicability across multiple NLU tasks. Such efforts would help better exploit the characteristics of the Vietnamese language and improve its performance in diverse NLU applications.

\subsection{Training Data Resource for Contrastive Learning}

Contrastive-like datasets are essential for training contrastive learning models, containing both positive (x, x+) and negative (x, x-) sentence pairs. Positive pairs represent semantically similar sentences, while negative pairs highlight dissimilarity, helping models learn to distinguish subtle semantic differences. These datasets are fundamental for tasks like natural language inference, where the goal is to determine the relationship between sentence pairs, such as entailment and contradiction. In English, datasets such as SNLI \cite{bowman2015large}, MultiNLI \cite{williams2018broad}, and ANLI \cite{nie2020adversarial} have been widely used for training and evaluating contrastive learning models \cite{gao2021simcse, wang2021cline}. Additionally, Chinese features the OCNLI dataset \cite{hu2020ocnli}, which has been successfully applied to train the CLOWER model \cite{chen2022clower}. For multilingual language NLI datasets, XNLI \cite{conneau2018xnli} and VietX-NLI \cite{bui2025vietx} provide a crucial resource for building multilingual contrastive learning models \cite{zhao2024leveraging, shuaibo2022supervised}, further extending the applicability of NLI.

For Vietnamese, recent years have witnessed the emergence of notable NLI datasets such as ViNLI \cite{van2022vinli}, ViANLI \cite{van2024vianli}, VnNewsNLI \cite{nguyen2022building}, and the dataset from the VLSP 2021 - vnNLI Challenge \cite{anh2022vlsp}. These datasets are considered valuable contrastive-like resources, offering significant potential for developing contrastive learning models and enhancing semantic representation in Vietnamese.

\section{Our Proposed Framework}
\label{sect:ViCSE}

\subsection{Overview}

\begin{figure*}[h!]
    \centering
    \includegraphics[width=1\linewidth]{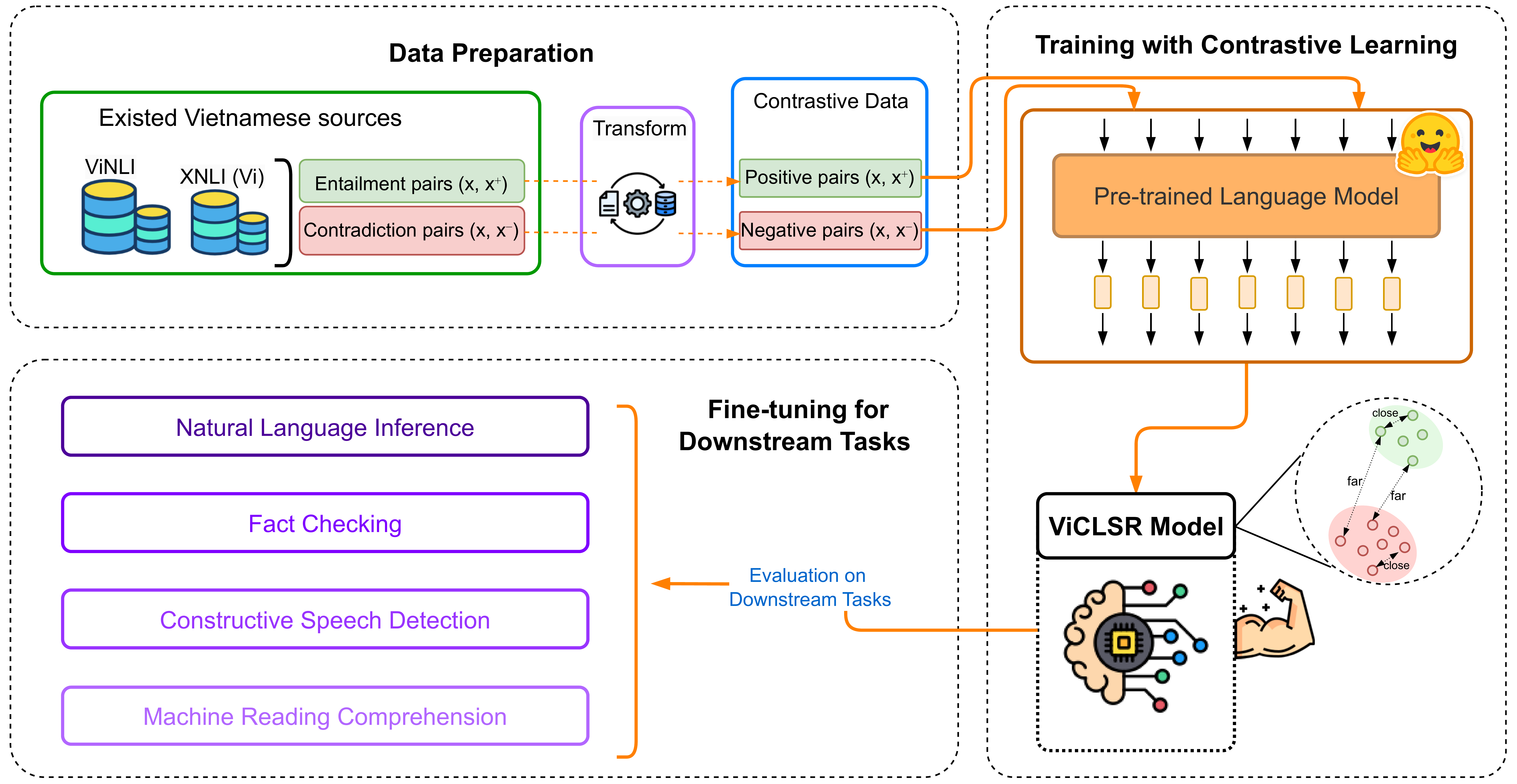}
    \caption{Overview of our supervised contrastive learning framework (ViCLSR) for Vietnamese NLU tasks, including data preparation, contrastive training, and fine-tuning for downstream tasks.}
    \label{fig:Overview}
\end{figure*}

An overview of our supervised contrastive learning framework, designed to improve sentence embeddings for Vietnamese text understanding, is presented in Figure \ref{fig:Overview}. This approach tackles the challenge of limited annotated data in low-resource languages like Vietnamese by leveraging contrastive learning techniques and adapting existing resources. By doing so, it enables the development of robust models that effectively capture semantic relationships between sentences and enhance performance across various Vietnamese NLU tasks.

The proposed framework comprises three main components: Data Preparation (Section \ref{sect:DataPreparation}), Training the Model with Contrastive Learning (Section \ref{sect:Training}), and Fine-tuning for Downstream Tasks (Section \ref{sect:Fine-tuning}).

\begin{itemize}
  \item \textbf{Data Preparation:} In this phase, available Vietnamese datasets are adapted for supervised contrastive learning. This involves utilizing labeled datasets, such as those from Vietnamese Natural Language Inference (NLI) tasks, to construct supervised contrastive data. Entailment pairs (\(x, x^+\)) are treated as positive examples, while contradiction pairs (\(x, x^-\)) serve as negative examples. This process ensures the availability of a substantial and high-quality dataset for contrastive learning.

  \item \textbf{Training with Contrastive Learning:} In this phase, a state-of-the-art pre-trained language model is selected and trained on the constructed contrastive dataset. The training process employs a supervised contrastive loss function, which minimizes the embedding distance between positive pairs and maximizes the distance between negative pairs. This ensures that the resulting embeddings capture the semantic nuances of Vietnamese text effectively.
  \item \textbf{Fine-tuning for Downstream Tasks:} The final stage involves adapting the trained contrastive model to specific downstream NLU tasks, such as natural language inference, fact checking, text classification, machine reading comprehension, and other Vietnamese language processing tasks. The fine-tuning process further optimizes the model for task-specific objectives, ensuring high performance and applicability.
\end{itemize}

This framework demonstrates the potential of supervised contrastive learning to overcome resource limitations in Vietnamese NLU, while also contributing to advancements in sentence representation learning for low-resource languages.

\subsection{Data Preparation}
\label{sect:DataPreparation}

\begin{algorithm*}[h!]
\caption{Data Preparation for Contrastive Learning}
\label{algorithm_Data_Preparation}
\begin{algorithmic}[1]
\State \textbf{Input:} 
\State \quad \textbf{NLI Dataset:} Premises \( P \), Hypotheses \( H \), Labels \( L \)

\State \textbf{Output:} 
\State \quad \textbf{Dataset:} Contrastive learning dataset based on NLI

\State Initialize empty datasets \( D_{NLI} \)

\State \textbf{Process NLI Dataset:}
\For{each \( (p, h, l) \in (P, H, L) \)}
    \If{ \( l = \text{entailment} \) }
        \State Assign \( p \to \text{sentence1} \), \( h \to \text{sentence2} \)
    \ElsIf{ \( l = \text{contradiction} \) }
        \State Assign \( h \to \text{hard\_neg} \)
    \EndIf
\EndFor
\State Construct \( D_{NLI} = \{ \text{sentence1 (\( x \))}, \text{sentence2 (\( x^{+} \))}, \text{hard\_neg (\( x^{-} \))} \} \)

\State \textbf{Return} \( D_{NLI} \)
\end{algorithmic}
\end{algorithm*}

\begin{table*}[ht]
\caption{Statistics of Data Generated for Contrastive Learning from NLI Datasets.}
\centering
\resizebox{\columnwidth}{!}{%
\begin{tabular}{lcccc}
\hline
\multirow{2}{*}{\textbf{Dataset}} & \multirow{2}{*}{\textbf{premise (x)}} & \multicolumn{2}{c}{\textbf{Hypothesis}} & \multirow{2}{*}{\textbf{(x, x+, x-)}} \\ \cline{3-4}
      &       & \multicolumn{1}{c}{\textbf{Entailment (x+)}} & \textbf{Contradiction (x-)} &       \\ \hline
ViNLI & 6,094 & \multicolumn{1}{c}{6,094}               & 6,094                  & 6,094 \\ 
XNLI  & 2,500 & \multicolumn{1}{c}{2,500}               & 2,500                  & 2,500 \\ \hline
Total & 8,594 & \multicolumn{1}{c}{8,594}               & 8,594                  & 8,594 \\ \hline
\end{tabular}%
}
\label{tab: Dataset_statistics}
\end{table*}

Inspired by SimCSE \cite{gao2021simcse}, we construct the contrastive training dataset directly from labeled Vietnamese NLI corpora (ViNLI \cite{van2022vinli} and XNLI (Vi) \cite{conneau2018xnli}) and approach this task in a supervised manner by transforming NLI labels into contrastive pairs. Specifically, we leverage the semantic relations already encoded in the entailment and contradiction labels to build positive and negative pairs, while neutral cases, which do not provide unambiguous similarity or dissimilarity cues, are excluded from the contrastive dataset to avoid introducing noise.

\begin{itemize}
    \item Positive pairs are derived from \textit{entailment} instances, where the premise and hypothesis express semantically consistent or semantically aligned meanings. This ensures that the encoder learns to place semantically similar sentences closer in the embedding space.
    \item Negative pairs are derived from \textit{contradiction} instances, where the hypothesis conflicts with the meaning of the premise. These pairs explicitly encourage the encoder to push apart semantically dissimilar sentences in the embedding space.
\end{itemize}

This label-based construction strategy provides high-quality contrastive pairs tailored to Vietnamese NLU. Importantly, to avoid any data leakage, we only use the training splits of ViNLI and XNLI (Vi) to construct contrastive pairs. The official dev and test splits of ViNLI are never included in contrastive pre-training and are reserved exclusively for downstream evaluation.

Algorithm \ref{algorithm_Data_Preparation} outlines the process of converting input data from NLI datasets into data appropriate for contrastive learning. Specifically, when using natural language inference datasets, we select the premise (\( x \)) and hypothesis (\( x^{+} \)) pairs with the entailment label as positive pairs (\( x \), \( x^{+} \)), and then select the hypothesis (\( x^{-} \)) with the contradiction label as the negative pair (\( x \), \( x^{-} \)) corresponding to the initial premise. The final output of the algorithm is a datasets for contrastive learning training: nli-based dataset. The statistics of the data we have for training contrastive learning are presented in Table \ref{tab: Dataset_statistics}.

\subsection{Training with Contrastive Learning}
\label{sect:Training}

\begin{figure*}[ht!]
    \centering
    \includegraphics[width=1\linewidth]{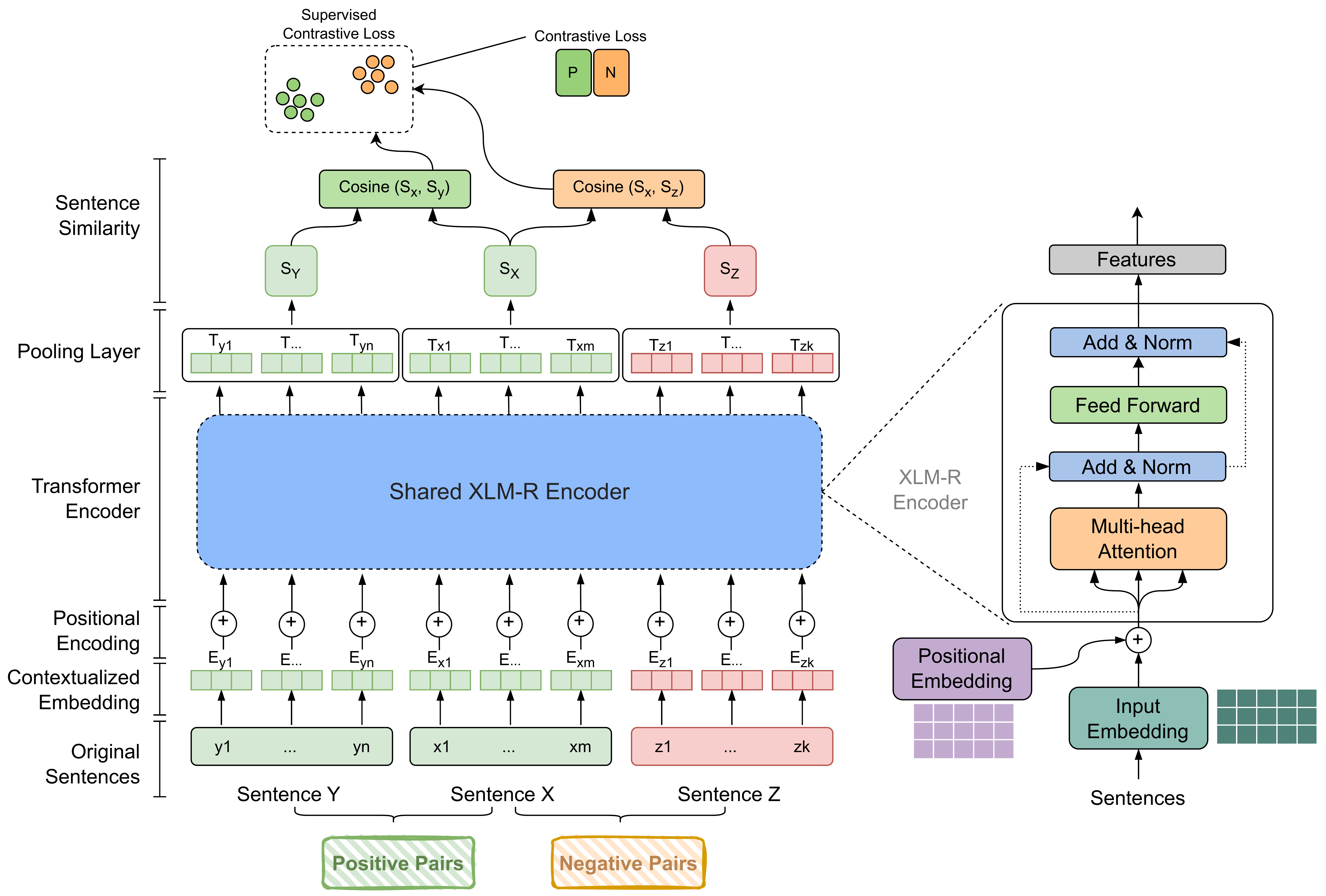}
    \caption{Our Contrastive Learning Architecture with a shared XLM-R encoder.}
    \label{fig:model}
\end{figure*}

Contrastive learning is a promising approach for training models to distinguish between semantically similar and dissimilar sentence pairs, which is crucial for generating high-quality sentence embeddings. However, low-resource languages like Vietnamese face significant challenges due to the limited availability of annotated data. Inspired by the supervised contrastive learning approach in SimCSE \cite{gao2021simcse}, which significantly improves sentence embeddings by leveraging natural language inference datasets as labeled pairs, we propose ViCLSR — a model that employs supervised CL to improve sentence embeddings specifically for Vietnamese.

Figure \ref{fig:model} illustrates the architecture of the ViCLSR model, which takes three sentences as input: one positive pair (Sentence X and Sentence Y: \( x \), \( x^{+} \)) and one negative pair (Sentence X and Sentence Z: \( x \), \( x^{-} \)). This architecture leverages the XLM-R encoder \cite{conneau-etal-2020-unsupervised}, a multilingual pre-trained model that has demonstrated strong performance across a wide range of NLU tasks and has supported 100 languages, making it easier to develop similar frameworks for other low-resource languages, and includes the following components.

\begin{itemize}
\item \textbf{Input Representation:} All three sentences (\(x, x^+, x^-\)) are first tokenized using the default \texttt{XLM-RobertaTokenizer} provided by HuggingFace. Each tokenized sentence is then transformed into input embeddings through the following steps:
\begin{itemize}
    \item \textbf{Token and Contextualized Embeddings:} Words are mapped to subword token embeddings by the XLM-R tokenizer, which are subsequently contextualized by the shared XLM-R encoder to capture their semantic meaning in context.
    \item \textbf{Positional Embeddings:} Learned positional embeddings are added to encode word order, which is crucial for representing sentence structure.
\end{itemize}
\end{itemize}

\begin{itemize}
    \item \textbf{Transformer Encoder:} The input embeddings are passed through a shared XLM-R encoder  which consists of multiple stacked transformer blocks. Each block follows the standard architecture with the following components:
    \begin{itemize}
        \item \textbf{Multi-head Attention}: This mechanism enables the model to attend to different parts of the input sequence simultaneously, capturing complex relationships between words.
        \item \textbf{Add \& Norm}: A residual connection and normalization step are applied to stabilize training.
        \item \textbf{Feed Forward Network}: A fully connected layer refines the output further.
        \item \textbf{Add \& Norm}: Another residual connection and normalization complete the transformer block.

    \end{itemize}
\end{itemize}

\begin{itemize}
    \item \textbf{Pooling Layer:} The output features from the encoder are aggregated into fixed-size vector representations for each sentence (\(S_x, S_y, S_z\)) using a pooling operation.
\end{itemize}

\begin{itemize}
    \item \textbf{Similarity Computation and Loss Function:}
    \begin{itemize}
        \item Measures similarity between the positive pair (\(S_x, S_y\)) and the negative pair (\(S_x, S_z\)).
         \item The model is trained using a supervised contrastive loss function, which minimizes the distance between embeddings of positive pairs while maximizing the distance between negative pairs. This ensures that the resulting sentence embeddings are well-aligned semantically. Similar to SimCSE \cite{gao2021simcse} and DiffCSE \cite{chuang2022diffcse}, in ViCLSR, we use a cosine similarity-based, temperature-scaled supervised contrastive loss to optimize the model. For a batch of training \( B = \{(x_i, x_i^+, x_i^-)\}_{i=1}^N \), where \( x_i \) is the premise sentence, \( x_i^+ \) is the entailment hypotheses sentence, and \( x_i^- \) is the contradiction hypotheses sentence. The supervised contrastive loss for the \(i\)-th training instance is formulated as Equation (\ref{equa-loss}).

\begin{equation}
L_i = -\log \left( \frac{ \exp \left( \text{sim}(h_i, h_i^+) / \tau \right) }{ \sum_{j=1}^{N} \exp \left( \text{sim}(h_i, h_j^+) / \tau \right) + \exp \left( \text{sim}(h_i, h_j^-) / \tau \right) } \right) .
\label{equa-loss}
\end{equation}

Here, $N$ represents the batch size, while $\mathbf{h}_i$, $\mathbf{h}_i^+$, and $\mathbf{h}_i^-$ denote the embeddings of the premise ($x$), the entailment hypothesis ($x^+$), and the contradiction hypothesis ($x^-$), respectively. These embeddings are generated by the XLM-R encoder, where $\mathbf{h}_i = f(x_i)$, $\mathbf{h}_i^+ = f(x_i^+)$, and $\mathbf{h}_i^- = f(x_i^-)$, with $f(\cdot)$ representing the encoder function. The similarity function $\text{sim}()$ is cosine similarity, which emphasizes the directional alignment between vectors. The temperature hyperparameter $\tau$ controls the sharpness of the similarity distribution, facilitating more effective differentiation between positive and negative pairs by scaling their similarity scores.

The similarity between two embeddings, $\mathbf{h}_1$ and $\mathbf{h}_2$, is measured using cosine similarity, defined as Equation (\ref{eq:cosine_similarity}).

\begin{equation}
\text{sim}(\mathbf{h}_1, \mathbf{h}_2) = \frac{\mathbf{h}_1 \cdot \mathbf{h}_2}{\|\mathbf{h}_1\| \|\mathbf{h}_2\|} .
\label{eq:cosine_similarity}
\end{equation}

This loss function encourages the model to minimize the distance between embeddings of positive pairs while maximizing the separation from negative pairs, enabling the model to learn robust and semantically meaningful sentence representations. It is particularly well-suited for tasks requiring nuanced understanding of semantic similarity, such as Vietnamese natural language inference.

    \end{itemize}
\end{itemize}

\subsection{Fine-tuning for Downstream NLU Tasks}

\begin{figure*}[ht]
    \centering
    \includegraphics[width=1\linewidth]{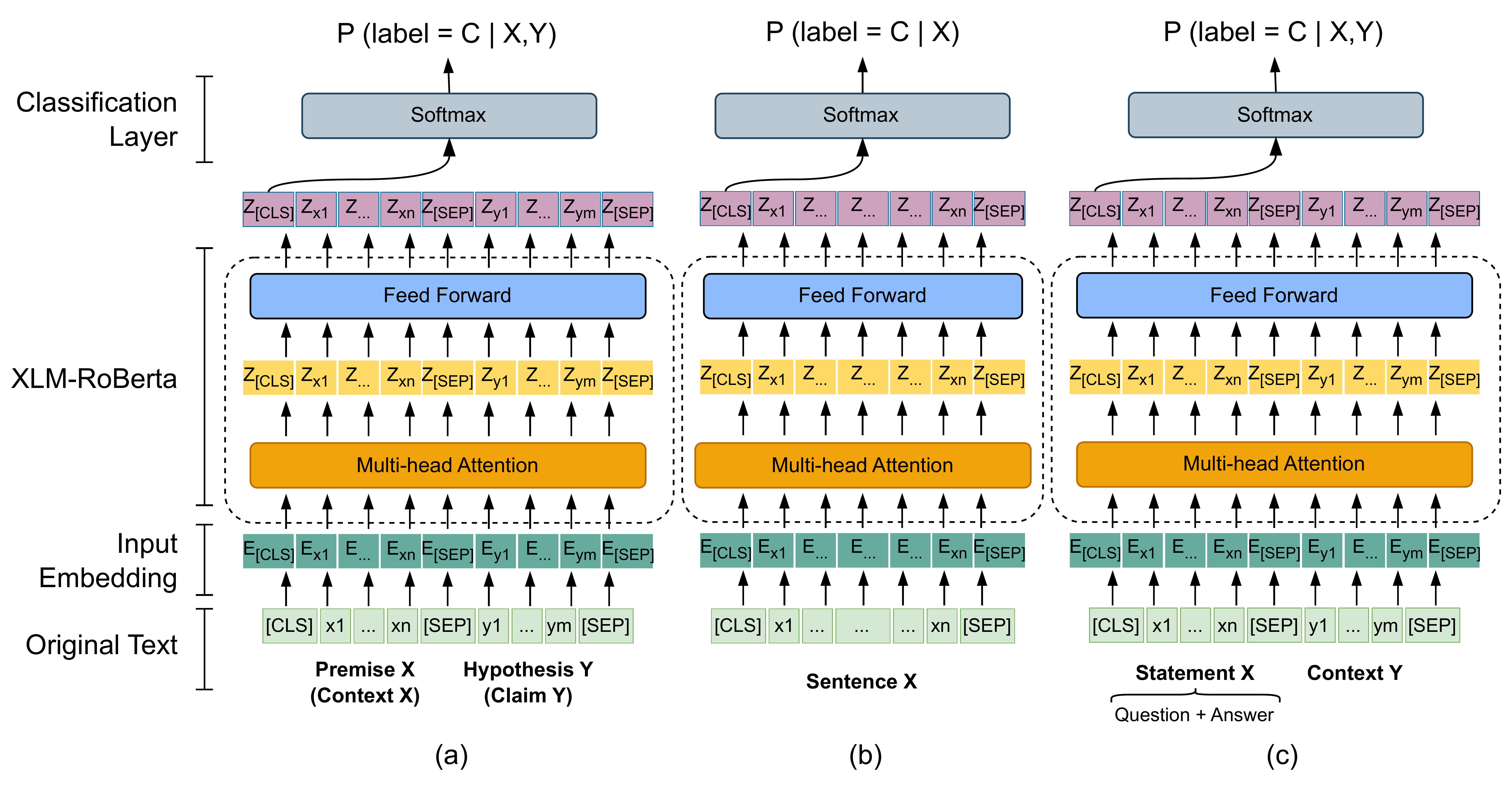}
    \caption{Architecture for Fine-tuning the ViCLSR Model on Various Downstream NLU Tasks. Figure 4(a) Represents Fine-tuning for Natural Language Inference and Fact Checking Tasks, Where the Model Predicts the Semantic Relationship Between a Premise (Context) and Hypothesis (Claim). Figure 4(b) Illustrates Fine-tuning for Constructive Speech Detection Tasks, Processing Single Input Sentences. Figure 4(c) Demonstrates Fine-tuning for Multiple-Choice Machine Reading Comprehension Tasks, Where the Model Predicts the Correct Answer Choice Based on the Given Context and Question.}
    \label{fig:fine-tune}
\end{figure*}

\label{sect:Fine-tuning}
Natural Language Understanding is a fundamental component that allows machines to interpret and process human language in a way that mirrors human cognitive abilities. It plays a crucial role in enhancing the performance of language models in NLU tasks, which require a profound understanding of context and meaning. The final stage of our approach involves adapting the contrastively trained ViCLSR model to various Vietnamese downstream NLU tasks, such as natural language inference, fact checking, constructive speech detection, and machine reading comprehension. This fine-tuning process optimizes the representations of the model to align with task-specific objectives, ensuring enhanced performance and applicability across Vietnamese language understanding tasks. Figure \ref{fig:fine-tune} illustrates the architecture for fine-tuning ViCLSR on these four downstream tasks described as follows.

\begin{itemize}
\item \textbf{Natural language inference and fact checking - Figure 4(a):} The model predicts the semantic relationship in NLI tasks (entailment, contradiction, or neutral) between two inputs: a \textit{Premise} and a \textit{Hypothesis}. Meanwhile, in fact checking, the model evaluates the veracity of a \textit{Claim} against a \textit{Context}, classifying the relationship as \textit{support}, \textit{refute}, or \textit{not-enough-information}. The process involves the following steps.
    \begin{itemize}
    \item \textbf{Input Transformation:} The \textit{Premise (Context)} - $X$ and \textit{Hypothesis (Claim)} - $Y$ are encoded with special tokens (\texttt{[CLS]} at the start and \texttt{[SEP]} between segments).
    \item \textbf{Embedding:} The text is transformed into input embeddings.
    \item \textbf{Encoding:} The embeddings are passed through the XLM-R model, which leverages \textit{multi-head attention} and a \textit{feed-forward network} for contextual representation.
    \item \textbf{Classification:} A classification layer with a softmax function predicts the probability of a label given the premise and hypothesis: \( P (label = C \| X, Y) \).
    \end{itemize}
\end{itemize}

\begin{itemize}
\item \textbf{Constructive speech detection - Figure 4(b):} For this tasks, where only a single input sentence ($X$) is used, the process is streamlined as follows.
    \begin{itemize}
    \item \textbf{Input Transformation:} The sentence is tokenized and encoded with \texttt{[CLS]} at the start and \texttt{[SEP]} at the end.
    \item \textbf{Embedding and Encoding:} The input is passed through the same pipeline as in Figure 3(a).
    \item \textbf{Classification:} The representation of the \texttt{[CLS]} token is fed into a classification layer with a softmax function to predict the probability of a label: \( P (label = C \| X) \)
    \end{itemize}
\end{itemize}

\begin{itemize}
\item \textbf{Machine reading comprehension - Figure 4(c):} The model is designed to predict the correct answer to a question based on a given \textit{context (Y)}. The process is as follows.
    \begin{itemize}
    \item \textbf{Input Transformation:} The Question and each answer choice are combined to form multiple \textit{Statements (X)}, where each Statement represents a potential answer. These Statements are then paired with the \textit{Context (Y)} for input into the model.
    \item \textbf{Embedding and Encoding:} Each \textit{Statement (X)} and \textit{Context (Y)} pair is passed through the XLM-R model to produce embeddings, similar to the NLI process.
    \item \textbf{Answer Prediction:} The model computes the probability distribution over the entailment label for each Statement (X) and Context (Y) pair. The label entailment corresponds to the correct answer, while contradiction corresponds to the incorrect answers. The Statement with the highest entailment probability is selected as the final answer.
    \end{itemize}
\end{itemize}


\section{Experiments and Results}
\label{sect:experiment}

\subsection{Benchmark Datasets}
\label{sub-sec-dataset}
To comprehensively evaluate the effectiveness of ViCLSR, we selected a diverse range of benchmark datasets that span multiple natural language processing tasks, including natural language inference (ViNLI \cite{van2022vinli}), fact checking (ViWikiFC \cite{le2024viwikifc} and ViFactCheck \cite{hoa2024vifactcheck}), constructive speech detection (UIT-ViCTSD \cite{nguyen2021constructive}), and machine reading comprehension (ViMMRC2.0 \cite{luu2023multiple}). Each dataset is aligned with the fine-tuning approach shown in Figure \ref{fig:fine-tune}, ensuring compatibility and allowing for a comprehensive evaluation of the core capabilities of ViCLSR. The information of the datasets is summarized in Table \ref{tab:NLP_task_exper}.

\begin{table*}[htbp]
\caption{Summary of the Benchmark Datasets and NLU Tasks Used to Evaluate the Performance of the ViCLSR Model.}
\centering
\resizebox{\textwidth}{!}{%
\begin{tabular}{lrrrllllc}
\hline
\textbf{Dataset} &
  \multicolumn{1}{c}{\textbf{Train}} &
  \multicolumn{1}{c}{\textbf{Dev}} &
  \multicolumn{1}{c}{\textbf{test}} &
  \multicolumn{1}{c}{\textbf{Task}} &
  \multicolumn{1}{c}{\textbf{Source}} &
  \multicolumn{1}{c}{\textbf{Domain}} &
  \multicolumn{1}{c}{\textbf{Metrics}} &
  \textbf{Classes} \\ \hline
\textbf{ViNLI}         & 24,376 & 3,009 & 2,991  & NLI                           & Online news       & Open domain & Acc/F1 & 3 \\
\textbf{ViWikiFC}      & 16,738 & 2,090 & 2,091  & Fact checking                 & Wikipedia         & Open domain & Acc/F1 & 3 \\
\textbf{ViFactCheck}      & 5,062 & 723 & 1,447  & Fact checking                 & Online news         & Open domain & F1 & 3 \\
\textbf{UIT-ViCTSD}    & 7,000  & 1,000 & 2,000  & Constructive speech detection & Social media      & Open domain & Acc/F1 & 2 \\
\textbf{ViMMRC2.0}     & 3,600  & 564   & 1,109  & Multiple-choice MRC           & Student textbooks & Literature  & Acc    & - \\ \hline
\end{tabular}%
}
\label{tab:NLP_task_exper}
\end{table*}

\textbf{ViNLI} \cite{van2022vinli} is a benchmark dataset for open-domain NLI research in Vietnam, focused on determining the logical relationship (entailment, contradiction, or neutral) between a premise and a hypothesis. It comprises over 30,000 human-annotated premise-hypothesis pairs, carefully collected from more than 800 online news articles spanning 13 distinct topics. To ensure balance and prevent topic bias, the dataset is evenly distributed across development and test sets. ViNLI serves as a rigorous benchmark for evaluating models’ ability to discern nuanced semantic relationships in open-domain contexts. The dataset consists of 24,376 training instances, 3,009 development instances, and 2,991 test instances.


\textbf{ViWikiFC} \cite{le2024viwikifc} is the first Vietnamese fact checking corpus derived from Wikipedia. It involves assessing the veracity of claims by analyzing their semantic relationship with context and classifying them as support, refute, or not enough information based on evidence from reliable sources. This dataset includes over 20,000 manually annotated claims, supported by evidence from diverse topics such as history, geography, philosophy, and science. Annotators followed strict guidelines to ensure quality and consistency, which makes ViWikiFC an essential resource for evaluating the ability of models to verify factual information. The dataset comprises 16,738 training instances, 2,090 development instances, and 2,091 test instances.

\textbf{ViFactCheck} \cite{hoa2024vifactcheck} is a publicly available benchmark dataset designed for multi-domain Vietnamese news fact checking. This dataset comprises 7,232 human-annotated claim-evidence pairs categorized into three labels: support, refute, or not enough information, similar to ViWikiFC. It is sourced from reputable Vietnamese online news outlets and covers 12 diverse topics. A notable feature of ViFactCheck is its focus on real-world fact checking scenarios, where annotators are required to generate intricate claims that combine multiple pieces of evidence. ViFactCheck serves as a valuable resource for evaluating the performance of state-of-the-art pre-trained language models and large language models in fact checking tasks. It is designed to test these models’ ability to process complex information, including identifying semantic ambiguities and navigating intricate inferential chains.

\textbf{UIT-ViCTSD} \cite{nguyen2021constructive} focuses on detecting constructive and toxic speech in Vietnamese social media comments. The dataset comprises 10,000 annotated comments categorized into labels such as “constructive,” “non-constructive,” “toxic,” and “non-toxic.” In our experiments with ViCLSR, we specifically targeted the task of determining whether a comment is constructive or not. This dataset serves as a valuable benchmark for evaluating the ability of models to promote healthier online discussions by effectively filtering non-constructive, toxic content. UIT-ViCTSD includes 7,000 training instances, 1,000 development instances, and 2,000 test instances.

\textbf{ViMMRC2.0} \cite{luu2023multiple} is a multiple-choice machine reading comprehension dataset designed to evaluate the ability of models to understand text by answering questions related to the content. This dataset is specifically created for educational purposes, containing 699 reading passages sourced from Vietnamese textbooks for grades 1 to 12. These passages, including both prose and poetry, are paired with 5,273 multiple-choice questions, with difficulty levels progressively increasing across grades. ViMMRC 2.0 serves as a valuable benchmark for evaluating language understanding in educational contexts. The dataset includes 3,600 training instances, 564 development instances, and 1,109 test instances.

\subsection{Comparative Models}
\begin{table*}[h!]
\caption{Summarizing the Detailed Information of Previous Pre-Trained Models Compared to Our Model (ViCLSR).}
\centering
\resizebox{\textwidth}{!}{%
\begin{tabular}{lrrrrrllc}
\hline
\textbf{Model} &
  \multicolumn{1}{c}{\textbf{\#Params}} &
  \multicolumn{1}{c}{\textbf{\#Layers}} &
  \multicolumn{1}{c}{\textbf{\#Heads}} &
  \multicolumn{1}{c}{\textbf{\#Hidden}} &
  \multicolumn{1}{c}{\textbf{\#Vocab}} &
  \multicolumn{1}{c}{\textbf{Language}} &
  \multicolumn{1}{c}{\textbf{\begin{tabular}[c]{@{}c@{}}Data \\ Source\end{tabular}}} &
  \textbf{\begin{tabular}[c]{@{}c@{}}CL\\ Strategy\end{tabular}} \\ \hline
\textbf{PhoBERT$_{Base}$}          & 135M & 12 & 12 & 768  & 64K  & Monolingual  & Wikipedia, News                                                & No  \\ 
\textbf{PhoBERT$_{Large}$}         & 370M & 24 & 16 & 1024 & 64K  & Monolingual  & Wikipedia, News                                                & No  \\ 
\textbf{mBERT}                & 179M & 12 & 12 & 768  & 119K & Multilingual & Wikipedia                                                      & No  \\ 
\textbf{XLM-R$_{Base}$}      & 270M & 12 & 8  & 768  & 250K & Multilingual & CommonCrawl                                                    & No  \\ 
\textbf{XLM-R$_{Large}$}             & 550M & 24 & 16 & 1024 & 250K & Multilingual & CommonCrawl                                                    & No  \\ 
\textbf{CafeBERT}             & 550M & 24 & 16 & 1024 & 250K & Multilingual & Wikipedia, News                                                    & No  \\
\textbf{DiffCSE}             &125M  &12  &12  &768  &50K  &Monolingual  &Wikipedia
  & Yes  \\

\hline
\textbf{ViCLSR} & 550M & 24 & 16 & 1024 & 250K & Multilingual & CommonCrawl, NLI & Yes \\ \hline
\end{tabular}%
}
\label{tab:detail-pre-trained}
\end{table*}

To evaluate the effectiveness of the ViCLSR model, we conducted a comprehensive experimental comparison with several state-of-the-art models, encompassing both monolingual and multilingual pre-trained models. These benchmarks, designed for a variety of NLU research tasks as presented in Section \ref{sub-sec-dataset}, help us determine the advantages of our proposed contrastive learning approach for Vietnamese text understanding. Table \ref{tab:detail-pre-trained} provides detailed parameter specifications for each model compared to our proposed model.

\textbf{PhoBERT} \cite{nguyen2020phobert}: This model represents the first large-scale pre-trained language model specifically tailored for Vietnamese, based on the RoBERTa architecture. It was trained on a 20GB corpus consisting of Vietnamese Wikipedia and curated news datasets. With its focus on the linguistic characteristics unique to Vietnamese, PhoBERT has demonstrated strong performance across various Vietnamese NLU tasks, establishing itself as a critical model for evaluating Vietnamese language models. The model is available in two versions, PhoBERT$_{Base}$ and PhoBERT$_{Large}$, both of which were evaluated in our experiments.

\textbf{CafeBERT} \cite{do2024vlue}: This model is a Vietnamese-adapted language model introduced alongside the VLUE benchmark for comprehensive Vietnamese NLU evaluation. It builds upon the multilingual XLM-R$_{Large}$ architecture through continued pretraining on an 18GB corpus of Vietnamese text, including approximately 1GB of Wikipedia and 17GB of curated news articles. The model retains the configuration of XLM-R$_{Large}$, consisting of 24 transformer layers, 16 attention heads, a hidden size of 1024, and roughly 550 million parameters. This adaptation enhances the linguistic proficiency of the model in Vietnamese, enabling it to outperform both monolingual and multilingual baselines across diverse NLU tasks. By combining cross-lingual generalization from XLM-R with domain-specific Vietnamese knowledge, CafeBERT achieves state-of-the-art performance on multiple datasets within the VLUE benchmark.

\textbf{Multilingual BERT} \cite{devlin2019bert}: Pre-trained on a massive corpus spanning 104 languages, including Vietnamese, mBERT provides strong cross-lingual transfer capabilities. With 12 layers, 12 attention heads, 768 hidden units, and approximately 179 million parameters, it is a versatile model for multilingual NLU tasks. Trained using a masked language modeling (MLM) objective, mBERT leverages a WordPiece vocabulary of approximately 119,000 tokens. However, due to its generalized nature, it may not fully capture the nuances of Vietnamese as effectively as monolingual models.

\textbf{XLM-R} \cite{conneau-etal-2020-unsupervised}: This model demonstrates exceptional capacity for language comprehension, surpassing both mBERT and PhoBERT in many tasks. Its superiority stems from its foundation in the robust RoBERTa architecture and its training on a massive multilingual dataset, CommonCrawl, encompassing 100 languages, including Vietnamese. XLM-R$_{Base}$ consists of 12 layers, 8 attention heads, and 270 million parameters, while XLM-R$_{Large}$ is more powerful, featuring 24 layers, 16 attention heads, and 550 million parameters. XLM-R has achieved outstanding results across a wide array of NLU tasks, making it an ideal benchmark for evaluating multilingual performance. We evaluate both XLM-R$_{Base}$ and XLM-R$_{Large}$ versions in our experiments.

\textbf{DiffCSE} \cite{chuang2022diffcse}: This model is an unsupervised contrastive learning framework that improves sentence embeddings through a difference-based objective. It introduces a difference prediction loss inspired by equivariant contrastive learning to better capture semantic variations. The model employs RoBERTa$_{Base}$ (12 layers, 12 attention heads, 768 hidden units, ~125M parameters) as the encoder. DiffCSE achieves state-of-the-art performance on semantic textual similarity and transfer tasks, establishing a strong unsupervised baseline for sentence representation learning.

\subsection{Experimental Settings}
\textbf{For training ViCLSR}, the model was fine-tuned using a supervised contrastive learning approach based on the XLM-R$_{Large}$ architecture. The training configuration included a learning rate of 1e-5, a train batch size of 32, and a total of 10 training epochs. The pooler type was set to CLS, ensuring that the contextualized embedding of the [CLS] token was utilized for generating sentence embeddings. These settings were carefully selected to maximize the performance of the model while ensuring stability during the fine-tuning process.

\textbf{For the fine-tuning of downstream tasks}, all models including mBERT, PhoBERT, XLM-R, CafeBERT, DiffCSE, and the proposed ViCLSR—were trained under the same experimental settings to ensure fairness and comparability. A learning rate of 3e-5 was applied to base models, while large models were fine-tuned with a learning rate of 1e-5. Other hyperparameters, such as a batch size of 16 and 7 training epochs, were consistently applied across models, with minor adjustments to accommodate task-specific characteristics. This standardized fine-tuning procedure enabled an effective and fair evaluation of model capabilities across various Vietnamese NLU tasks.

To ensure a robust and comprehensive evaluation of all models on downstream tasks, a standardized experimental setup was employed. All experiments were conducted within a Google Colaboratory\footnote{https://colab.research.google.com/} environment, utilizing the computational power of NVIDIA L4 GPUs. The implementation leveraged the Transformers library provided by Huggingface\footnote{https://huggingface.co/}, which offers pre-trained models and fine-tuning utilities, streamlining the experimentation process. This consistent environment ensured comparability and reproducibility in all model assessments.

\subsection{Evaluation Metrics}
To evaluate the performance of ViCLSR and other models on the downstream tasks. For the tasks of natural language inference, fact checking, and constructive speech detection, the evaluation metrics of \textbf{Accuracy} and \textbf{macro-averaged F1-score} were applied following the evaluation methodologies established in the corresponding datasets: ViNLI \cite{van2022vinli}  for NLI, ViWikiFC \cite{le2024viwikifc} and ViFactCheck \cite{hoa2024vifactcheck} for fact checking, and UIT-ViCTSD \cite{nguyen2021constructive} for constructive speech detection. By adopting these standard practices, a consistent and reliable comparison of model performance across tasks was ensured.

\begin{itemize}
    \item \textbf{Accuracy}: Accuracy measures the proportion of correctly classified instances among the total number of instances and is calculated as Equation (\ref{eq:accuracy}).
    \begin{equation}
        \text{Accuracy} = \frac{\text{Number of Correct Predictions}}{\text{Total Number of Instances}} .
        \label{eq:accuracy}
    \end{equation}
    
    \item \textbf{Macro-averaged F1-Score}: Macro-averaged F1-score provides a balanced measure of the model performance by combining precision and recall, considering all classes equally. The formula is presented as Equation (\ref{eq:f1score_nli}).
    \begin{equation}
        \text{F1} = 2 \times \frac{\text{Precision} \times \text{Recall}}{\text{Precision} + \text{Recall}} .
        \label{eq:f1score_nli}
    \end{equation}
    Where Precision and Recall are defined as Equation (\ref{eq:precision_nli}) and Equation (\ref{eq:recall_nli}) respectively.
    \begin{equation}
        \text{Precision} = \frac{\text{True Positives}}{\text{True Positives} + \text{False Positives}} .
        \label{eq:precision_nli}
    \end{equation}
    \begin{equation}
        \text{Recall} = \frac{\text{True Positives}}{\text{True Positives} + \text{False Negatives}} .
        \label{eq:recall_nli}
    \end{equation}
\end{itemize}

For the Multiple-Choice MRC task, we evaluated the models using the \textbf{Accuracy} (Acc) metric, following the approach used in the ViMMRC2.0 dataset \cite{luu2023multiple}. This metric calculates the proportion of questions for which the model selects the correct answer from the answer options. The formula is presented as Equation (\ref{eq:accuracy_mc_mrc}).
    \begin{equation}
        \text{Acc} = \frac{\text{Number of Correct Answers}}{\text{Total Number of Questions}} .
        \label{eq:accuracy_mc_mrc}
    \end{equation}

\subsection{Experimental Results}

\label{sect:main_result}
\begin{table*}[h!]
\caption{The Result of Model Performance with Our ViCLSR Across NLU Tasks.}
\centering
\resizebox{\textwidth}{!}{%
\begin{tabular}{l|cc|ccc|cc|c}
\hline
\multirow{3}{*}{\textbf{Models}} &
  \multicolumn{2}{c|}{\textbf{NLI}} &
  \multicolumn{3}{c|}{\textbf{Fact Checking}} &
  \multicolumn{2}{c|}{\textbf{\begin{tabular}[c]{@{}c@{}}Constructive Speech\\  Detection\end{tabular}}} &
  \textbf{MRC} \\ \cline{2-9} 
 &
  \multicolumn{2}{c|}{\textbf{ViNLI}} &
  \multicolumn{2}{c|}{\textbf{ViWikiFC}} &
  \textbf{ViFactCheck} &
  \multicolumn{2}{c|}{\textbf{UIT-ViCTSD}} &
  \textbf{ViMMRC2.0} \\ \cline{2-9} 
 &
  \multicolumn{1}{c}{\textbf{Acc}} &
  \textbf{F1} &
  \multicolumn{1}{c}{\textbf{Acc}} &
  \multicolumn{1}{c|}{\textbf{F1}} &
  \textbf{F1} &
  \multicolumn{1}{c}{\textbf{Acc}} &
  \textbf{F1} &
  \textbf{Acc} \\ \hline
PhoBERT$_{Base}$ &
  \multicolumn{1}{c}{72.87} &
  72.79 &
  \multicolumn{1}{c}{80.67} &
  \multicolumn{1}{c|}{80.70} &
  77.76 &
  \multicolumn{1}{c}{79.40} &
  78.59 &
  53.92 \\ 
PhoBERT$_{Large}$ &
  \multicolumn{1}{c}{75.93} &
  75.87 &
  \multicolumn{1}{c}{81.63} &
  \multicolumn{1}{c|}{81.62} &
  79.76 &
  \multicolumn{1}{c}{80.70} &
  76.86 &
  54.73 \\ 
mBERT &
  \multicolumn{1}{c}{64.84} &
  64.83 &
  \multicolumn{1}{c}{75.93} &
  \multicolumn{1}{c|}{76.01} &
  69.94 &
  \multicolumn{1}{c}{78.90} &
  76.70 &
  53.38 \\ 
XLM-R$_{Base}$ &
  \multicolumn{1}{c}{71.59} &
  71.51 &
  \multicolumn{1}{c}{78.53} &
  \multicolumn{1}{c|}{78.56} &
  81.10 &
  \multicolumn{1}{c}{80.40} &
  78.10 & 
  \multicolumn{1}{c}{29.39}\\ 
XLM-R$_{Large}$ &
  \multicolumn{1}{c}{81.36} &
  81.31 &
  \multicolumn{1}{c}{85.11} &
  \multicolumn{1}{c|}{85.15} &
  88.02 &
  \multicolumn{1}{c}{81.70} &
  79.44 &
  57.52 \\ 
CafeBERT &
  \multicolumn{1}{c}{82.06} & 82.05
   &
  \multicolumn{1}{c}{85.22} &
  \multicolumn{1}{c|}{85.24} & 87.45
   &
  \multicolumn{1}{c}{83.00} &81.53
   &57.98
   \\
DiffCSE &
  \multicolumn{1}{c}{58.35} &58.34
   &
  \multicolumn{1}{c}{66.14} &
  \multicolumn{1}{c|}{66.12} &61.84
   &
  \multicolumn{1}{c}{81.80} &80.36
   &41.11
   \\
  \hline
\textbf{ViCLSR} &
  \multicolumn{1}{c}{\textbf{82.86}} &
  \textbf{82.84} &
  \multicolumn{1}{c}{\textbf{86.56}} &
  \multicolumn{1}{c|}{\textbf{86.57}} &
  \textbf{88.78} &
  \multicolumn{1}{c}{\textbf{83.40}} &
  \textbf{82.22} &
  \textbf{59.06} \\ \hline
\textbf{Compared to XLM-R} &
  \multicolumn{1}{c}{\textcolor{blue}{$\uparrow$1.50}} &
  \textcolor{blue}{$\uparrow$1.53} &
  \multicolumn{1}{c}{\textcolor{blue}{$\uparrow$1.45}} &
  \multicolumn{1}{c|}{\textcolor{blue}{$\uparrow$1.42}} &
  \textcolor{blue}{$\uparrow$0.76} &
  \multicolumn{1}{c}{\textcolor{blue}{$\uparrow$1.70}} &
  \textcolor{blue}{$\uparrow$2.78} &
  \textcolor{blue}{$\uparrow$1.54} \\ \hline

  \textbf{Compared to PhoBERT} &
  \multicolumn{1}{c}{\textcolor{blue}{$\uparrow$6.93}} &
  \textcolor{blue}{$\uparrow$6.97} &
  \multicolumn{1}{c}{\textcolor{blue}{$\uparrow$4.93}} &
  \multicolumn{1}{c|}{\textcolor{blue}{$\uparrow$4.97}} &
  \textcolor{blue}{$\uparrow$9.02} &
  \multicolumn{1}{c}{\textcolor{blue}{$\uparrow$2.70}} &
  \textcolor{blue}{$\uparrow$5.36} &
  \textcolor{blue}{$\uparrow$4.33} \\ \hline
  \textbf{Compared to CafeBERT} &
  \multicolumn{1}{c}{\textcolor{blue}{$\uparrow$0.80}} &
  \textcolor{blue}{$\uparrow$0.79} &
  \multicolumn{1}{c}{\textcolor{blue}{$\uparrow$1.34}} &
  \multicolumn{1}{c|}{\textcolor{blue}{$\uparrow$1.33}} &
  \textcolor{blue}{$\uparrow$1.33} &
  \multicolumn{1}{c}{\textcolor{blue}{$\uparrow$0.40}} &
  \textcolor{blue}{$\uparrow$0.69} &
  \textcolor{blue}{$\uparrow$1.08} \\ \hline
  \textbf{Compared to DiffCSE} &
  \multicolumn{1}{c}{\textcolor{blue}{$\uparrow$24.51}} &
  \textcolor{blue}{$\uparrow$24.50} &
  \multicolumn{1}{c}{\textcolor{blue}{$\uparrow$20.42}} &
  \multicolumn{1}{c|}{\textcolor{blue}{$\uparrow$20.45}} &
  \textcolor{blue}{$\uparrow$26.94} &
  \multicolumn{1}{c}{\textcolor{blue}{$\uparrow$2.35}} &
  \textcolor{blue}{$\uparrow$1.86} &
  \textcolor{blue}{$\uparrow$17.95} \\ \hline
\end{tabular}%
}
\label{tab:main-results}
\end{table*}

The experimental results of ViCLSR and several competitive baselines are summarized in Table \ref{tab:main-results}. Overall, ViCLSR achieves superior performance across four representative Vietnamese NLU tasks — natural language inference, fact checking, constructive speech detection, and multiple-choice machine reading comprehension. The model consistently surpasses both monolingual models such as PhoBERT and CafeBERT, and multilingual counterparts including mBERT and XLM-R, as well as a recent contrastive baseline, DiffCSE. These results demonstrate that our supervised contrastive learning framework effectively enhances Vietnamese sentence representations, leading to stronger generalization across diverse downstream understanding tasks.

For natural language inference, ViCLSR achieved an accuracy of 82.86\% and a macro-averaged F1-score of 82.84\% on the ViNLI dataset, establishing the highest performance among all evaluated models. Compared with the strongest multilingual baseline, XLM-R$_{Large}$, ViCLSR achieved improvements of about 1.50\% in accuracy and 1.50\% in F1-score. The gap became considerably larger when compared with the monolingual PhoBERT$_{Large}$, where ViCLSR performed roughly 7\% better in accuracy, confirming the effectiveness of contrastive supervision over conventional encoder-only architectures. The model also slightly outperformed the Vietnamese-adapted CafeBERT by approximately 0.8\% accuracy and showed a substantial margin over the contrastive baseline DiffCSE, exceeding it by more than 24\% in both accuracy and F1-score. These consistent improvements demonstrate that the proposed supervised contrastive learning framework captures richer and more discriminative semantic representations, enabling ViCLSR to model fine-grained sentence relationships and generalize effectively across diverse inference patterns.

For fact checking, ViCLSR consistently outperformed both multilingual and monolingual baselines across the ViWikiFC and ViFactCheck datasets. The model achieved notable gains over XLM-R$_{Large}$ and CafeBERT, while maintaining a clear performance margin compared to PhoBERT$_{Large}$ and the contrastive DiffCSE model. Specifically, on ViWikiFC, ViCLSR surpassed PhoBERT$_{Large}$ by more than 4\% F1-score and outperformed DiffCSE by over 20\% F1-score; on ViFactCheck, the improvements were even larger over 9\% F1-score against PhoBERT$_{Large}$ and more than 26\% F1-score compared to DiffCSE. These substantial gains underscore the superior ability of ViCLSR to capture factual consistency and detect misleading or contradictory claims. Overall, the results confirm that the proposed supervised contrastive framework not only enhances semantic discrimination but also strengthens factual reasoning in Vietnamese text understanding.

For constructive speech detection, ViCLSR achieved strong results on the UIT-ViCTSD dataset, reaching 83.40\% accuracy and 82.22\% F1-score. The model achieved a slight improvement over CafeBERT, but demonstrated a much larger advantage compared to PhoBERT$_{Large}$, with nearly 3\% higher accuracy, and maintained a clear lead of more than 2\% accuracy over the contrastive DiffCSE model. These findings confirm that ViCLSR effectively captures the pragmatic and semantic nuances that distinguish constructive from non-constructive speech. The results further underscore the model’s ability to represent discourse-level meaning, which is essential for promoting healthier and more responsible online communication.

For machine reading comprehension, ViCLSR achieved 59.06\% accuracy on the ViMMRC 2.0 dataset, showing a moderate improvement over all other models. Our model exhibited a more notable gain compared to PhoBERT$_{Large}$, with approximately 4\% higher accuracy, and maintained a substantial advantage of over 17\% accuracy against the contrastive DiffCSE baseline. Although the overall accuracy is lower than in other tasks due to the inherent complexity of multi-sentence reasoning, these results demonstrate that ViCLSR possesses strong comprehension and reasoning capabilities, effectively generalizing across challenging question-answering contexts.

The ViCLSR model effectively demonstrates the advantages of incorporating supervised contrastive learning to enhance Vietnamese sentence representations, achieving state-of-the-art performance across a wide range of NLU tasks. The model consistently surpasses both monolingual encoders, such as PhoBERT and CafeBERT, and multilingual baselines, including mBERT and XLM-R, while also outperforming the contrastive DiffCSE model by a substantial margin. These consistent improvements highlight the robustness and generalization capability of ViCLSR, reinforcing its potential as an effective and scalable framework for low-resource languages like Vietnamese.

\hl{Furthermore, to provide empirical support regarding model convergence stability and explicitly address potential overfitting risks on smaller datasets like ViMMRC2.0, we present the training and validation loss curves for all evaluated tasks in Appendix} \ref{loss_curve}. \hl{These loss curves confirm that selecting the best model based on validation scores effectively prevents overfitting, ensuring stable performance across all evaluated tasks.}

\section{Results Analysis and Discussion}
\label{sect:result_analysis}
In this section, we conduct a comprehensive analysis of the ViCLSR model to better understand its language understanding capabilities. We perform hyperparameter and ablation studies to optimize the configuration of the model (see Section \ref{lab:hyper}), followed by an evaluation of its semantic representation capability and the distribution of sentence embeddings (see Section \ref{lab:Semantic-Representation-Capability} and Section \ref{lab:Distribution}). Next, we present a sensitivity analysis that examines the effect of different training data sizes on the performance of ViCLSR (see Section \ref{training_size}). In addition, we analyze the attention mechanism to explore how the model handles sentence relationships (see Section \ref{attention}). Finally, we provide recommendations for adapting the model to other low-resource languages (see Section \ref{recommend}). These analyses aim to offer valuable insights into the strengths and limitations of ViCLSR across various aspects of natural language understanding.

\subsection{Hyperparameter and Ablation Studies}
\label{lab:hyper}

\textbf{Temperature Hyperparameter $\tau$:}  To evaluate the impact of the temperature hyperparameter on the performance of the proposed ViCLSR model, we investigate its effect in the contrastive loss function. The temperature controls the sharpness of the probability distribution in the contrastive learning objective, influencing how the model differentiates between similar and dissimilar sentence pairs. \hl{Following the experimental protocol of SimCSE, we treat $\tau$ as a fixed global scaling parameter and determine its value through grid search on the development set, keeping it constant during pretraining to enable controlled analysis of its effect.} In this experiment, ViCLSR was trained with multiple $\tau$ values ranging from 0.001 to 1 and subsequently fine-tuned on five downstream NLU tasks: ViNLI, ViWikiFC, ViFactCheck, UIT-ViCTSD, and ViMMRC2.0.

\begin{table}[htbp]
\centering
\begin{tabular}{lrrrrrr}
\hline
\multicolumn{1}{c}{$\tau$} &
  \multicolumn{1}{c}{0.001} &
  \multicolumn{1}{c}{0.01} &
  \multicolumn{1}{c}{0.05} &
  \multicolumn{1}{c}{0.1} &
  \multicolumn{1}{c}{0.5} &
  \multicolumn{1}{c}{1} \\ \hline
ViNLI       & 82.66 & 82.48 & \textbf{83.28} & 82.04 & 82.75 & 81.46 \\ 
ViWikiFC    & 87.70 & 86.65 & \textbf{88.56} & 86.79 & 86.75 & 87.03 \\ 
ViFactCheck & \textbf{88.38} & 88.11 & 87.97 & 87.00 & 87.23 & 85.20 \\ 
UIT-ViCTSD  & 81.59 & 82.30 & \textbf{82.75} & 82.35 & 80.55 & 82.70 \\ 
ViMMRC2.0   & 55.67       & 55.62      & \textbf{56.38} & 55.85      & 55.11       & 55.02      \\ \hline
\end{tabular}%
\caption{Impact of Temperature Hyperparameter $\tau$ in Contrastive Loss on Fine-Tuning Accuracy for Downstream NLU Tasks.}
\label{tab:Temperature}
\end{table}

The results in Table \ref{tab:Temperature} present the development set accuracy of ViCLSR across downstream tasks when fine-tuned with different values of $\tau$. The findings indicate that the choice of the temperature hyperparameter has a significant effect on model accuracy across various NLU tasks. The best overall performance is observed at $\tau$ = 0.05, achieving the highest accuracy on most datasets, such as 83.28\% on ViNLI, 88.56\% on ViWikiFC, 82.75\% on UIT-ViCTSD, and 56.38\% on ViMMRC2.0. These results suggest that $\tau$ = 0.05 provides the best balance in the contrastive loss function, producing well-calibrated similarity scores.

Deviating from this optimal value leads to performance degradation. When $\tau$ is too small (e.g., 0.001), the model accuracy drops due to overly sharp probability distributions, as seen in ViNLI (82.66\%) and ViWikiFC (87.70\%). Conversely, when $\tau$ is too large (e.g., 1.0), the accuracy also declines (81.46\% on ViNLI and 55.02\% on ViMMRC2.0), likely because a flatter probability distribution reduces contrastive learning effectiveness. Interestingly, ViCLSR achieves its highest accuracy on the ViFactCheck dataset when trained with $\tau$ = 0.001, reaching 88.38\%. This observation suggests that a sharper contrastive distribution can occasionally benefit datasets that require finer-grained semantic discrimination. Overall, these analyses reinforce the importance of properly tuning the temperature hyperparameter to maximize sentence representation quality and overall model performance.

\hl{Recent studies have further explored dynamic or adaptive temperature scheduling strategies to enhance contrastive learning optimization} \cite{manna2025dynamically}, \hl{where the temperature is adjusted during training to account for evolving embedding distributions or varying negative sample hardness. Such approaches have shown potential in improving convergence stability and boundary refinement, particularly in unsupervised or large-scale heterogeneous settings. In contrast, our framework adopts a fixed temperature following SimCSE} \cite{gao2021simcse} \hl{and DiffCSE} \cite{chuang2022diffcse}\hl{, as our supervised NLI-based setup already provides structured and semantically grounded contradiction pairs that serve as informative hard negatives. Nonetheless, our sensitivity analysis across a wide range of $\tau$ values offers insight into how temperature scaling affects representation geometry in this supervised context. Incorporating adaptive temperature scheduling into ViCLSR remains a promising extension that could further refine training dynamics without altering the core supervised contrastive formulation.}

\textbf{Ablation Tests on Auxiliary MLM Objective:} To investigate whether incorporating the auxiliary Masked Language Modeling (MLM) objective improves the performance of ViCLSR on Vietnamese NLU tasks, we conducted an ablation study to assess its effect during training. The weighting parameter $\lambda$ controls the influence of the MLM objective on the overall loss, allowing us to balance the contributions of the contrastive and MLM objectives, similar to the approach of Gao et al. (2021) \cite{gao2021simcse}, where the overall loss in Eq. \ref{equa-loss} is formulated as $L = L_{CL} + \lambda L_{MLM}$. In this experiment, ViCLSR was trained with multiple $\lambda$ values (0.001, 0.01, 0.05, 0.1, 0.5, and 1) and compared with a baseline model trained without the MLM objective (w/o MLM). The models obtained from different $\lambda$ settings were then fine-tuned and evaluated on downstream NLU tasks, including ViNLI, ViWikiFC, ViFactCheck, UIT-ViCTSD, and ViMMRC2.0.

The results in Table \ref{tab:Auxiliary-MLM} present the development set accuracy of ViCLSR across these downstream tasks under different $\lambda$ configurations. The findings show that excluding the MLM objective (w/o MLM) produces the highest accuracy on most tasks, such as 83.28\% on ViNLI and 88.56\% on ViWikiFC. When the MLM objective is included, model performance generally declines across datasets, except for ViFactCheck, where ViCLSR achieves its best result at $\lambda$ = 0.05. This observation suggests that while the MLM objective can enhance token-level representations, it may also interfere with the contrastive objective optimized for sentence-level semantic alignment. Consequently, excluding the MLM objective is more effective for tasks focusing on sentence-level understanding, whereas its inclusion should be applied selectively for datasets that benefit from richer token-level contextualization.

\begin{table}[htbp]
\centering
\begin{tabular}{lccccccc}
\hline
\multicolumn{1}{c}{$\lambda$} & w/o            & 0.001 & 0.01  & 0.05  & 0.1   & 0.5   & 1     \\ \hline
ViNLI                            & \textbf{83.28} & 82.08 & 83.10 & 82.04 & 82.57 & 82.35 & 82.44 \\ 
ViWikiFC                         & \textbf{88.56} & 86.41 & 86.99 & 88.18 & 86.89 & 87.99 & 87.13 \\ 
ViFactCheck                      & 87.97 & 87.11 & 87.28 & \textbf{88.11} & 86.86 & 87.55 & 86.58 \\ 
UIT-ViCTSD                       & \textbf{82.75} & 81.69 & 82.30 & 82.49 & 82.23 & 82.15 & 80.40 \\ 
ViMMRC2.0                        & \textbf{56.38} & 54.28       & 55.27      & 55.73      & 54.78      & 54.78       & 54.28      \\ \hline
\end{tabular}%

\caption{Effect of Auxiliary MLM Objective on Downstream NLU Tasks Performance Under Different $\lambda$ Settings.}
\label{tab:Auxiliary-MLM}
\end{table}

\textbf{Masking Rate:} To further explore the impact of the auxiliary MLM objective and address the concern regarding masking configurations, we conducted an additional analysis focusing on masking-rate variations under a fixed masking paradigm. In the previous experiments, the weighting parameter $\lambda$ controlled the contribution of the MLM objective to the overall loss. As discussed earlier, the configuration without the MLM objective (w/o MLM) achieved the highest overall performance, while a small contribution of the MLM loss ($\lambda$ = 0.05) produced relatively competitive results on several tasks such as ViWikiFC, ViFactCheck, and UIT-ViCTSD.

\begin{figure*}[ht]
    \centering
    \includegraphics[width=1\linewidth]{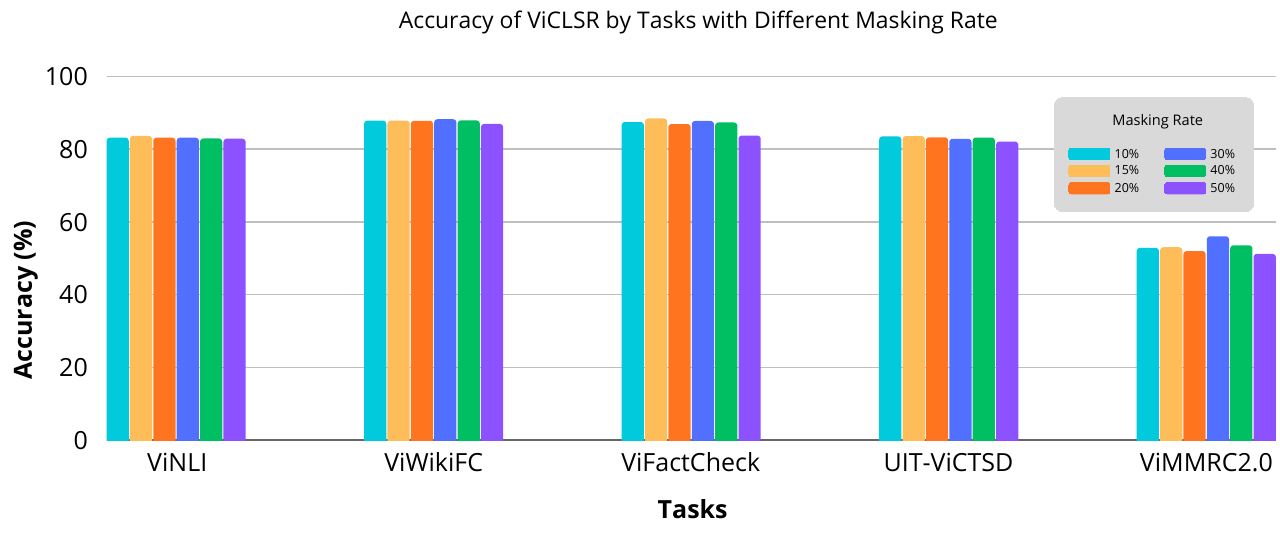}
    \caption{Impact of masking rate during contrastive learning on the downstream performance of ViCLSR across five Vietnamese NLU tasks.}
    \label{fig:masking-rate}
\end{figure*}

\hl{Importantly, our backbone model, XLM-R, already employs dynamic subword-level masking, where mask positions are re-sampled across epochs. This dynamic mechanism introduces stochastic regularization and exposes the model to diverse contextual patterns without requiring architectural modification. While alternative masking paradigms (e.g., whole-word or span masking) alter the structural granularity of token corruption, our objective is to isolate the interaction between contrastive supervision and contextual reconstruction under a controlled setting.

Therefore, rather than modifying the masking paradigm itself, we systematically vary its strength.} Specifically, under the configuration that includes the MLM objective ($\lambda$ = 0.05), we trained ViCLSR with masking rates of 10\%, 15\%, 20\%, 30\%, 40\%, and 50\%, and compared them against the commonly used default rate of 15\% \cite{devlin2019bert}. After pretraining under each masking configuration, the resulting ViCLSR models were fine-tuned and evaluated on five downstream NLU tasks: ViNLI, ViWikiFC, ViFactCheck, UIT-ViCTSD, and ViMMRC2.0. The development set accuracies of these configurations are summarized in Figure \ref{fig:masking-rate}.

Overall, the results in Figure \ref{fig:masking-rate} show that the performance of ViCLSR remains relatively stable across moderate masking rates, with the optimal configuration varying slightly across tasks. The commonly used masking rate of 15\% yields the highest accuracy for ViNLI, ViFactCheck, and UIT-ViCTSD, whereas a higher masking rate of 30\% performs best for ViWikiFC and ViMMRC2.0.  Very low masking rates (e.g., 10\%) \hl{provide limited contextual perturbation, which may reduce the diversity of reconstruction signals and limit abstraction}. Conversely, excessively high masking rates (e.g., 50\%) \hl{remove too much contextual information, which may hinder stable sentence-level abstraction}. These findings indicate that the masking rate strongly influences the balance between token-level reconstruction and sentence-level abstraction, and that maintaining a moderate range of 15\%–30\% allows ViCLSR to achieve stable and well-balanced performance across Vietnamese NLU tasks.

\textbf{Pooling Strategies: } Besides, we evaluated the impact of different pooling strategies on the effectiveness of sentence representations in ViCLSR. Pooling methods aggregate token-level embeddings into a single sentence-level representation, which significantly influences the performance of sentence embeddings. We tested four pooling methods: {[}CLS] Pooling (using embeddings of {[}CLS] token), Mean Pooling (averaging embeddings of all tokens), First-Last Pooling (averaging embeddings from the first and last layers), and Top-2 Pooling (averaging embeddings from the top two layers). The model was fine-tuned on the ViNLI, ViWikiFC, ViFactCheck, UIT-ViCTSD, and ViMMRC2.0 datasets for each strategy, and the fine-tuning accuracy on dev set was recorded in Table \ref{tab:Pooling-type}. 

\begin{table}[htbp]
\centering
\begin{tabular}{lcccc}
\hline
\multicolumn{1}{c}{\textbf{Pooling Type}} & {[}CLS{]} & Mean & First-Last & Top-2 \\ \hline
ViNLI       & \textbf{83.28} & 82.93 & 82.66 & 82.50 \\ 
ViWikiFC    & \textbf{88.56} & 87.37 & 87.51 & 87.56 \\ 
ViFactCheck & \textbf{87.97} & 87.00 & 87.14 & 87.00 \\ 
UIT-ViCTSD  & \textbf{82.75} & 82.64      & 82.24      & 82.02      \\ 
ViMMRC2.0   & \textbf{56.38} & 55.49       & 55.16      & 55.92      \\ \hline
\end{tabular}%
\caption{Impact of Different Pooling Strategies on Downstream NLU Tasks.}
\label{tab:Pooling-type}
\end{table}

The [CLS] Pooling method consistently achieves the highest accuracy among all evaluated strategies, highlighting its effectiveness in capturing global sentence-level semantics. Specifically, it outperforms all other methods across datasets, achieving 83.28\% on ViNLI, 88.56\% on ViWikiFC, and 87.97\% on ViFactCheck, demonstrating its robustness across different tasks. The results show that using the [CLS] token is more effective than averaging across multiple layers.

In contrast, Mean Pooling, which averages all token embeddings, results in a slight performance drop, with scores of 82.93\% on ViNLI and 87.37\% on ViWikiFC, likely due to feature dilution. First-Last Pooling and Top-2 Pooling produce competitive but suboptimal results,  showing that deeper representations do not always lead to better accuracy. These findings emphasize the critical role of pooling strategies in optimizing sentence embeddings, as an ineffective method can limit model performance even with strong pre-trained representations. [CLS] Pooling remains the most effective, reinforcing the importance of selecting a strategy that preserves key semantic structures for downstream NLU tasks.



\subsection{Semantic Representation Capability}
\label{lab:Semantic-Representation-Capability}

To evaluate the semantic representation capability of the ViCLSR model, which is trained using contrastive learning to optimize for capturing sentence similarity and dissimilarity, we compared our model with other baseline models on the information retrieval (IR) task. The evaluation was conducted on the fact checking dataset (ViWikiFC), where each claim sentence is paired with one most relevant evidence sentence in the context to verify its factuality.

For this experiment, we filtered the dataset to include only samples with the support label and computed accuracy across four scenarios: Top 1, Top 3, Top 5, and Top 10 most similar sentences predicted by the model. We evaluate according to Accuracy@TopK as shown in  Equation (\ref{eq:accuracy_topk}).

\begin{equation}
\label{eq:accuracy_topk}
\text{Accuracy@TopK} = \frac{1}{n} \sum_{i=1}^{n} \mathbb{I} \left( y_i \in \text{TopK}(x_i, C) \right) .
\end{equation}

\noindent where:

\begin{itemize}
    \item $n$: The total number of claim sentences in the dataset.
    \item $x_i$: The $i$-th claim sentence.
    \item $y_i$: The corresponding evidence sentence for the claim $x_i$.
    \item $C$: The set of context sentences.
    \item $\text{TopK}(x_i, C)$: The TopK set of sentences that the model predicts as the most similar to $x_i$ in $C$, based on cosine similarity.
    \item $\mathbb{I}(\cdot)$: The indicator function, which returns:
    \begin{itemize}
        \item $1$, if $y_i \in \text{TopK}(x_i, C)$;
        \item $0$, otherwise.
    \end{itemize}
\end{itemize}

To obtain sentence representations, we employed two pooling strategies: [CLS] pooling and Mean pooling. Table \ref{tab:IR-Acc} compares the retrieval accuracy of our model, ViCLSR, against several strong baselines, including XLM-R$_{Large}$, PhoBERT$_{Large}$, mBERT, CafeBERT, and the contrastive learning model DiffCSE, across different Accuracy@TopK settings. The results show that our proposed model, ViCLSR, achieves the best retrieval performance, significantly surpassing both multilingual models (mBERT and XLM-R$_{Large}$), monolingual Vietnamese models (PhoBERT$_{Large}$ and CafeBERT), and the contrastive learning baseline DiffCSE across all four TopK scenarios and both pooling strategies.
Notably, ViCLSR demonstrates exceptional accuracy at the top ranks (Top 1 and Top 3), highlighting its strength in prioritizing the most relevant evidence, which is crucial for IR tasks. Furthermore, while broader ranges like Top 10 indicate that there is potential to further improve the ranking quality, ViCLSR consistently retrieves relevant sentences effectively, indicating its robustness and practical applicability in retrieval scenarios where comprehensive evidence coverage is required.

\begin{table*}[h]
\caption{Comparison of IR System Accuracy Between ViCLSR and Other Models on ViWikiFC Using {[}CLS{]} Pooling and Mean Pooling.}
\centering
\begin{tabular}{llcccc}
\hline
\textbf{Pooling Type}              & \textbf{Models} & \textbf{Top 1}  & \textbf{Top 3}  & \textbf{Top 5}  & \textbf{Top 10} \\ \hline
\multirow{4}{*}{{[}CLS{]} Pooling} & ViCLSR       & \textbf{82.62} & \textbf{90.96} & \textbf{93.50} & \textbf{96.04} \\ 
 & XLM-R$_{Large}$     & 16.52 & 19.06 & 19.35 & 21.18 \\  
 & PhoBERT$_{Large}$ & 24.43 & 28.38 & 29.51 & 31.92 \\  
 & mBERT         & 48.72 & 59.32 & 63.55 & 67.65 \\
& CafeBERT &42.37  &50.00  &54.94  &59.60  \\  
 & DiffCSE         &40.68  &49.01  &52.54  &56.50  \\
 \hline
\multirow{4}{*}{Mean Pooling}      & ViCLSR       & \textbf{83.05} & \textbf{92.51} & \textbf{94.35} & \textbf{97.32} \\  
 & XLM-R$_{Large}$     & 54.24 & 61.02 & 64.69 & 67.09 \\  
 & PhoBERT$_{Large}$ & 76.69 & 86.02 & 87.99 & 90.82 \\  
 & mBERT         & 78.25 & 86.86 & 89.41 & 92.23 \\ 
 & CafeBERT &72.60  &78.95  &81.36  &84.89  \\  
 & DiffCSE         &44.07  &52.82  &56.92  &62.43  \\
 \hline
\end{tabular}%
\label{tab:IR-Acc}
\end{table*}

The differing performance of [CLS] pooling and Mean pooling in ViNLI (Table \ref{tab:Pooling-type}) and IR (Table \ref{tab:IR-Acc}) tasks reflects the distinct requirements of these tasks. For ViNLI, which focuses on sentence-level semantic relationships, the [CLS] token captures global semantics, making it more effective. In contrast, IR requires retrieving evidence from a large pool of context sentences, where Mean pooling aggregates information from all tokens, ensuring that relevant tokens contribute to the representation. This broader coverage makes Mean pooling better suited for IR tasks.

\subsection{Distribution of Sentence Embeddings}
\label{lab:Distribution}

We analyze the distribution of sentence embeddings generated by the ViCLSR model (ViCLSR) on the ViNLI dataset. The analysis focuses on two primary metrics: Alignment refers to how closely the embeddings of semantically similar pairs are grouped together and how effectively the embeddings of semantically dissimilar pairs are pushed apart, while Uniformity evaluates the extent to which embeddings are evenly distributed across the hypersphere, ensuring they are not overly clustered or concentrated in specific regions. The goal of this analysis is to evaluate the quality of embeddings produced by the model and compare it against baseline models, including XLM-R$_{Large}$, PhoBERT$_{Large}$, mBERT, CafeBERT, and DiffCSE. These metrics provide insights into the ability of the models to encode semantic relationships effectively and maintain generalizable embedding distributions.

\begin{figure*}[]
\centering
  \begin{subfigure}{.40\textwidth}
  \centering
    \includegraphics[width=1\linewidth]{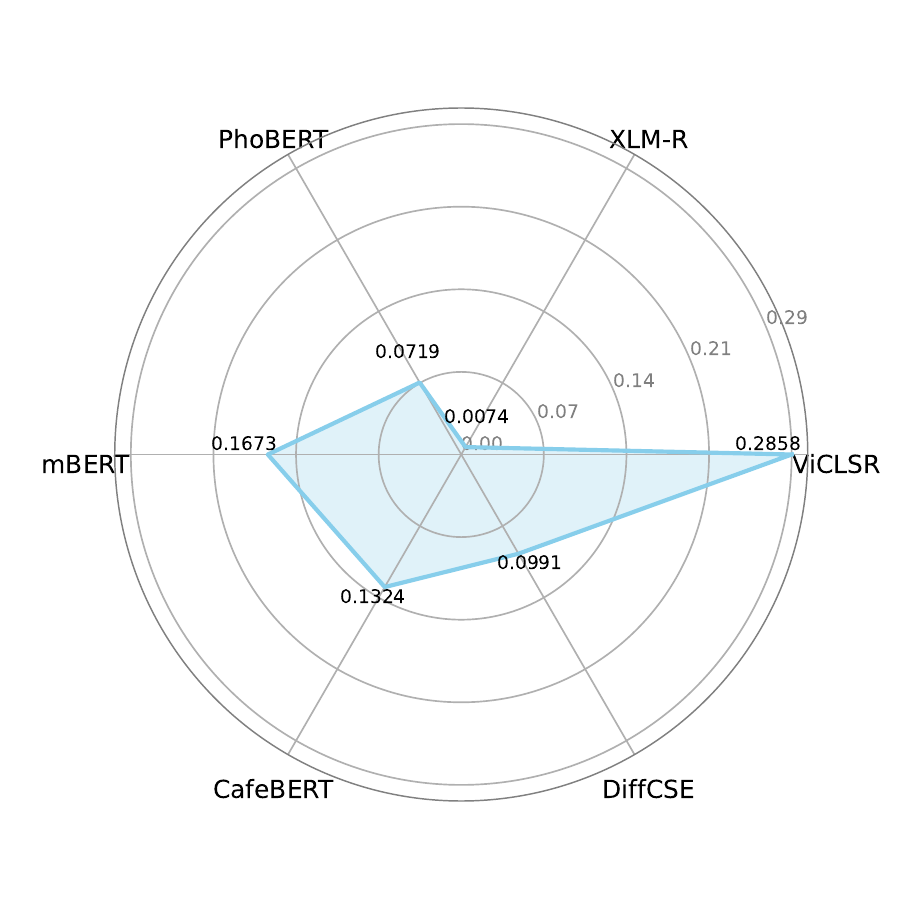}
    \caption{Alignment - E}
    \label{fig:Alignment-E}
  \end{subfigure}%
  \begin{subfigure}{.40\textwidth}
  \centering
\includegraphics[width=1\linewidth]{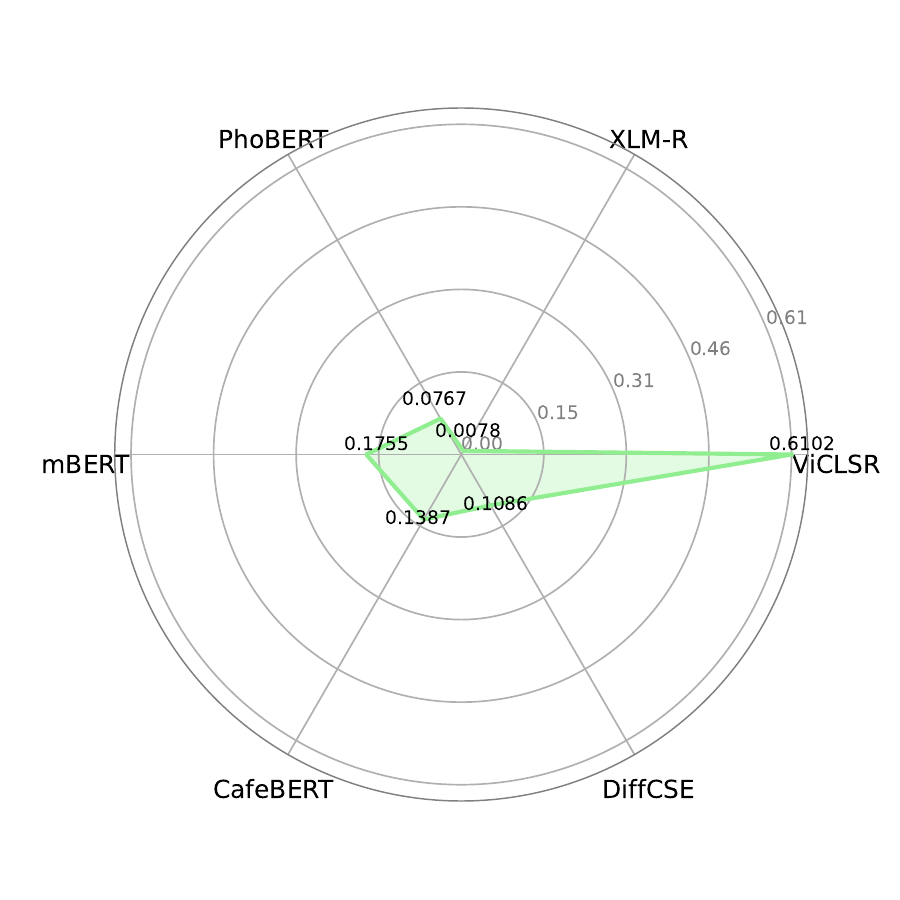}
    \caption{Alignment - C}
    \label{fig:Alignment-C}
  \end{subfigure}
    \begin{subfigure}{.40\textwidth}
  \centering
    \includegraphics[width=1\linewidth]{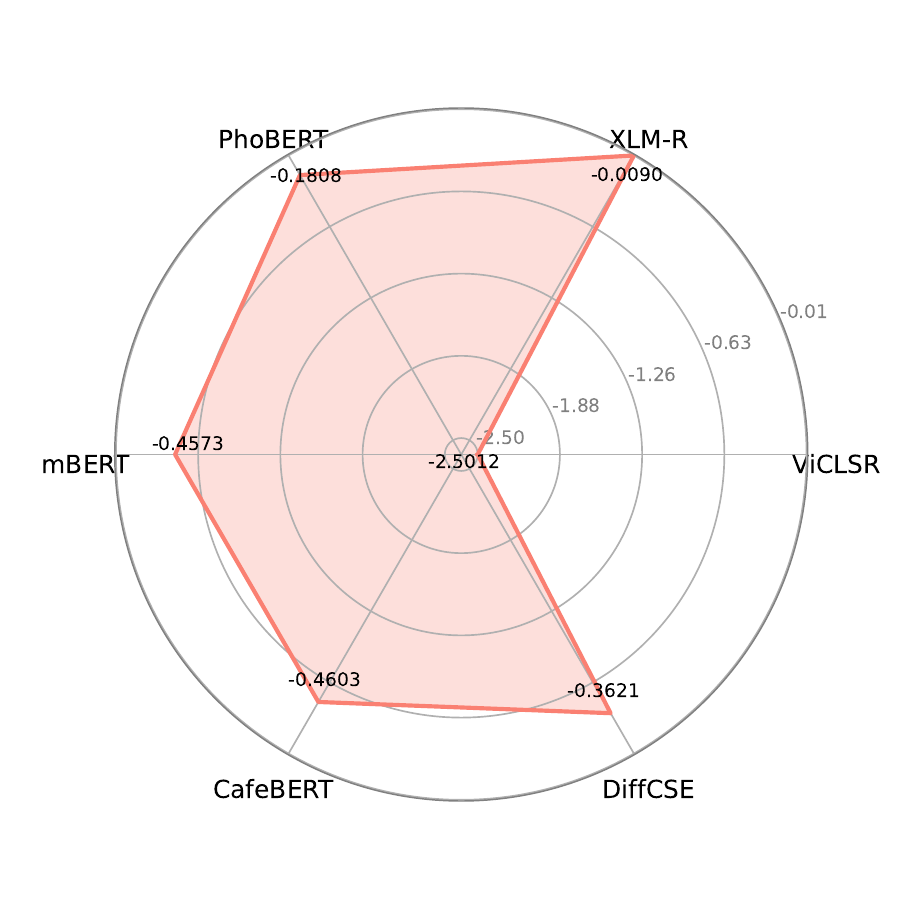}
    \caption{Uniformity}
    \label{fig:Uniformity}
  \end{subfigure}%
  \caption{Analysis of Alignment and Uniformity on ViNLI Dataset Comparing ViCLSR with Other Models Using Sentence Embeddings Derived from CLS Pooling (E: Alignment for Positive Pairs, C: Alignment for Negative Pairs).}
  \label{fig:Alignment-Uniformity}
\end{figure*}

The Alignment metric was calculated as the squared distance between the embeddings of semantically similar pairs (E) and dissimilar pairs (C). Specifically, positive pairs correspond to sentence pairs with an entailment relationship, while negative pairs correspond to sentence pairs with a contradiction relationship in the ViNLI dataset. Alignment is calculated as Equation (\ref{equa-align}).

\begin{equation}
\ell_{\text{align}} \triangleq \mathbb{E}_{(x, x^+)\sim p_{\text{pos}}} \|f(x) - f(x^+)\|^2 .
\label{equa-align}
\end{equation}

The Uniformity metric was calculated to evaluate how well the embeddings are distributed uniformly across the hypersphere. This computation was performed on all sentence pairs present in the ViNLI dataset. Uniformity is calculated as Equation (\ref{equa-uniform}).

\begin{equation}
\ell_{\text{uniform}} \triangleq \log \mathbb{E}_{x, y \sim p_{\text{data}}} e^{-2\|f(x) - f(y)\|^2} .
\label{equa-uniform}
\end{equation}


Figure \ref{fig:Alignment-C} shows that ViCLSR achieves an Alignment - C score of 0.6102, significantly outperforming all other models. This demonstrates the exceptional ability of the model to separate semantically dissimilar sentence pairs (e.g., contradiction pairs). This high score indicates that ViCLSR is highly effective in pushing embeddings of contradictory sentences apart, which is crucial to accurately capture semantic differences in sentence relationships.

With an Uniformity score of -2.5012 of the ViCLSR model as shown in Figure \ref{fig:Uniformity} shows that our model outperforms the baseline models by a wide margin. This indicates that the embeddings produced by the model are evenly distributed across the hypersphere, avoiding over-clustering or collapsing into dense regions. This distribution ensures the generalizability of the model and prevents it from being biased towards specific regions of the embedding space. In contrast, other models such as XLM-R$_{Large}$ (-0.0090),  PhoBERT$_{Large}$ (-0.1808), DiffCSE (-0.3621), mBERT (-0.4573), and  CafeBERT (-0.4603) show much poorer Uniformity scores, suggesting that their embeddings are less evenly distributed and may suffer from over-concentration, which hinders their ability to generalize across diverse semantic contexts.

Although ViCLSR performs as expected on Alignment E, with a score of 0.2858 (indicating good grouping of similar sentence pairs) as shown in Figure \ref{fig:Alignment-E}, it is noticeable that other models, such as XLM-R$_{Large}$, PhoBERT$_{Large}$, and DiffCSE exhibit unusually low scores. This suggests a phenomenon of “false low Alignment - E”, where embeddings of semantically similar pairs are not adequately grouped. This behavior, combined with similar Alignment - E and Alignment - C values in these models, points to a possible issue: over-concentration of embeddings in the hypersphere. Such over-concentration is supported by the extremely low Uniformity scores of these models, indicating that their embeddings collapse into dense regions, leading to poor differentiation between similar and dissimilar sentences. 

\begin{longtable}{|p{3.5cm}|p{3cm}|p{4cm}|}
\caption{Comparison of Retrieved Top-3 Examples by ViCLSR Against XLM-R and PhoBERT on the ViWikiFC Dataset. Retrieved sentences highlighted in \textcolor{BlueGreen}{blue-green text} represent correct retrievals, whereas those in \textcolor{Bittersweet}{bittersweet text} denote incorrect retrievals.}\label{tab:Top3-IR}
\\ \hline
\multicolumn{1}{|c|}{\textbf{ViCLSR}} & 
\multicolumn{1}{c|}{\textbf{XLM-R$_{Large}$}} & 
\multicolumn{1}{c|}{\textbf{PhoBERT$_{Large}$}} \\ 
\hline
\endfirsthead

\hline
\multicolumn{1}{|c|}{\textbf{ViCLSR}} & 
\multicolumn{1}{c|}{\textbf{XLM-R$_{Large}$}} & 
\multicolumn{1}{c|}{\textbf{PhoBERT$_{Large}$}} \\ 
\hline
\endhead

\multicolumn{3}{|p{11cm}|}{\textbf{Query 1}: "\textcolor{orange}{\textbf{Tô Định là kẻ vô cùng xấu xa và tàn độc.}}" \textit{(Tô Định was an extremely wicked and cruel man.)}

\textbf{Gold evidence}: "\textcolor{blue}{\textbf{Năm 39, thái thú quận Giao Chỉ là Tô Định tàn ác, giết chồng của Trưng Trắc là Thi Sách.}}" \textit{(In the year 39, Tô Định, the cruel governor of Giao Chỉ district, killed Thi Sách, the husband of Trưng Trắc.)}
} \\ \hline

\textbf{\#1.} \textcolor{BlueGreen}{Năm 39, thái thú quận Giao Chỉ là Tô Định tàn ác, giết chồng của Trưng Trắc là Thi Sách.}
\textit{(In the year 39, Tô Định, the cruel governor of Giao Chỉ district, killed Thi Sách, the husband of Trưng Trắc.)}
&
\textbf{\#1.} \textcolor{Bittersweet}{Nhà Thương thường phái quân đội đi chiến đấu chống lại những bộ tộc lân cận.} 
\textit{(The Shang dynasty often sent its armies to fight against neighboring tribes.)}
& 
\textbf{\#1.} \textcolor{Bittersweet}{Điều này hiển nhiên rất khó khiến người ta tin tưởng.} 
\textit{(This is obviously very hard for people to believe.)} \\ \hline

\textbf{\#2.} \textcolor{Bittersweet}{Năm Cam, Khánh Trắng là một số ví dụ về băng nhóm tội phạm có tổ chức.} 
\textit{(Năm Cam and Khánh Trắng are examples of organized criminal gangs.)} &
\textbf{\#2.} \textcolor{Bittersweet}{Các trang phục của người dân tộc thiểu số cũng có thể sử dụng.}
\textit{(The traditional costumes of ethnic minorities can also be used.)} 
& 
\textbf{\#2.} \textcolor{Bittersweet}{Theo cư dân địa phương, bà rất linh thiêng.} 
\textit{(According to local residents, she is considered very sacred.)} \\ \hline

\textbf{\#3.} \textcolor{Bittersweet}{Năm 931, Dương Đình Nghệ là tướng cũ của Khúc Hạo đem quân đánh phủ thành Đại La.} 
\textit{(In 931, Dương Đình Nghệ, a former general of Khúc Hạo, led his army to attack the Đại La citadel.)} &
\textbf{\#3.} \textcolor{Bittersweet}{Chính quyền này kiểm soát chặt chẽ mọi mặt của đời sống.} 
\textit{(This government tightly controls all aspects of life.)}
& 
\textbf{\#3.} \textcolor{Bittersweet}{Chính quyền này kiểm soát chặt chẽ mọi mặt của đời sống.} 
\textit{(This government tightly controls all aspects of life.)} \\ \hline

\multicolumn{3}{|p{11cm}|}{\textbf{Query 2}: "\textcolor{orange}{\textbf{Singapore sở hữu 12 khu vực công nghiệp.}}" \textit{(Singapore has 12 industrial zones.)}

\textbf{Gold\_evidence}: "\textcolor{blue}{\textbf{Singapore có 12 khu vực công nghiệp lớn, trong đó lớn nhất là Khu công nghiệp Jurong.}}" \textit{(Singapore has 12 major industrial zones, the largest of which is the Jurong Industrial Estate.)}
} \\ \hline

\textbf{\#1.} \textcolor{BlueGreen}{Singapore có 12 khu vực công nghiệp lớn, trong đó lớn nhất là Khu công nghiệp Jurong.} 
\textit{(Singapore has 12 major industrial zones, the largest of which is the Jurong Industrial Estate.)}
&
\textbf{\#1.} \textcolor{Bittersweet}{Tại đó, người dân bầu cho mọi việc.} 
\textit{(At that place, the people vote on everything.)}
& 
\textbf{\#1.} \textcolor{Bittersweet}{Kinh tế của Palawan chủ yếu phụ thuộc vào nông nghiệp.}  
\textit{(Palawan’s economy mainly depends on agriculture.)} \\ \hline

\textbf{\#2.} \textcolor{Bittersweet}{Singapore có cơ sở hạ tầng và một số ngành công nghiệp phát triển cao hàng đầu châu Á.} 
\textit{(Singapore has world-class infrastructure and some of the most advanced industries in Asia.)} &
\textbf{\#2.} \textcolor{Bittersweet}{Các phương tiện có động cơ không được tham gia lưu thông.} 
\textit{(Motor vehicles are not allowed to operate on the roads.)}
& 
\textbf{\#2.} \textcolor{Bittersweet}{Hệ thống giao thông công chánh ở Singapore rất phát triển.} 
\textit{(Singapore’s public transportation system is highly developed.)} \\ \hline

\textbf{\#3.} \textcolor{Bittersweet}{Quảng Nam có Khu kinh tế mở Chu Lai nổi tiếng với nhà máy của THACO.} 
\textit{(Quang Nam has the Chu Lai Open Economic Zone, which is famous for the THACO automobile plant.)} &
\textbf{\#3.} \textcolor{Bittersweet}{Và thực tế đúng là như vậy.} 
\textit{(And that is indeed true.)}
& 
\textbf{\#3.} \textcolor{Bittersweet}{Các công ty có liên kết với chính phủ kiểm soát hầu hết truyền thông nội địa tại Singapore.} 
\textit{(Government-linked companies control most of the domestic media in Singapore.)} \\ \hline

\multicolumn{3}{|p{11cm}|}{\textbf{Query 3}: "\textcolor{orange}{\textbf{Vùng duyên hải có mức thủy triều trung bình. }}" \textit{(The coastal zone has an average tidal range.)}

\textbf{Gold\_evidence}: "\textcolor{blue}{\textbf{Vùng duyên hải bao phủ khu vực nằm giữa các mức thủy triều cao và thấp nhất, nó là khu vực chuyển tiếp giữa các điều kiện đại dương và đất liền.}}" \textit{(The coastal zone covers the area between the highest and lowest tide levels; it is a transitional region between oceanic and terrestrial conditions.)}
} \\ \hline

\textbf{\#1.} \textcolor{Bittersweet}{Các dòng biển phần nào đó điều hòa khí hậu.} 
\textit{(Ocean currents help regulate the climate to some extent.)}
&
\textbf{\#1.} \textcolor{Bittersweet}{1942: Hiến chương Đại Tây Dương được ký kết.} 
\textit{(1942: The Atlantic Charter was signed.)}
& 
\textbf{\#1.} \textcolor{Bittersweet}{Các khu vực nhỏ hơn của đại dương được gọi là các biển, vịnh hay một số các tên gọi khác.} 
\textit{(Smaller areas of the ocean are called seas, bays, or by other names.)} \\ \hline

\textbf{\#2.} \textcolor{Bittersweet}{Các khu vực nhỏ hơn của đại dương được gọi là các biển, vịnh hay một số các tên gọi khác.} 
\textit{(Smaller areas of the ocean are called seas, bays, or by other names.)} &
\textbf{\#2.} \textcolor{Bittersweet}{Các dòng biển phần nào đó điều hòa khí hậu.} 
\textit{(Ocean currents help regulate the climate to some extent.)}
& 
\textbf{\#2.} \textcolor{Bittersweet}{Nhờ nằm giữa biển Đông nên quần đảo Hoàng Sa có khí hậu điều hòa, không quá lạnh về mùa đông, không quá nóng về mùa hè nếu so với những vùng đất cùng vĩ độ trong lục địa.} 
\textit{(Thanks to its location in the middle of the East Sea, the Paracel Islands enjoy a moderate climate — not too cold in winter and not too hot in summer compared to inland areas at the same latitude.)} \\ \hline

\textbf{\#3.} \textcolor{Bittersweet}{Nhờ nằm giữa biển Đông nên quần đảo Hoàng Sa có khí hậu điều hòa, không quá lạnh về mùa đông, không quá nóng về mùa hè nếu so với những vùng đất cùng vĩ độ trong lục địa.} 
\textit{(Thanks to its location in the middle of the East Sea, the Paracel Islands enjoy a moderate climate — not too cold in winter and not too hot in summer compared to inland areas at the same latitude.)} &
\textbf{\#3.} \textcolor{Bittersweet}{Các khu vực nhỏ hơn của đại dương được gọi là các biển, vịnh hay một số các tên gọi khác.} 
\textit{(Smaller areas of the ocean are called seas, bays, or by other names.)}
& 
\textbf{\#3.} \textcolor{Bittersweet}{1942: Hiến chương Đại Tây Dương được ký kết.}
\textit{(1942: The Atlantic Charter was signed.)} \\ \hline

\multicolumn{3}{|p{11cm}|}{\textbf{Query 4}: "\textcolor{orange}{\textbf{Nó có độ nóng chảy ở mức gần 30 độ C.}}" \textit{(It has a melting point of around 30°C.)}

\textbf{Gold\_evidence}: "\textcolor{blue}{\textbf{Nó là một kim loại kiềm mềm, màu bạc, và với điểm nóng chảy là 28 °C (83 °F) khiến cho nó trở thành một trong các kim loại ở dạng lỏng tại hay gần nhiệt độ phòng.}}" \textit{(It is a soft, silvery alkali metal, and with a melting point of 28°C (83°F), it is one of the few metals that are liquid at or near room temperature.)}
} \\ \hline

\textbf{\#1.} \textcolor{Bittersweet}{Ở trung tâm của Trái Đất, nhiệt độ có thể đạt tới 7000K và áp suất có thể lên tới 360 Gpa.} 
\textit{(At the Earth’s core, temperatures can reach up to 7000 K and pressures can rise to 360 GPa.)}
&
\textbf{\#1.} \textcolor{Bittersweet}{Đường phố được dành cho người đi bộ thưởng lãm.} 
\textit{(The streets are reserved for pedestrians to stroll and enjoy the scenery.)}
& 
\textbf{\#1.} \textcolor{Bittersweet}{Nếu tình trạng tiến triển thành say nắng, thì da nóng, khô là điển hình  khi các mạch máu giãn ra trong nỗ lực tăng mất nhiệt.} 
\textit{(If the condition progresses to heatstroke, hot and dry skin is typical as blood vessels dilate in an effort to increase heat loss.)} \\ \hline

\textbf{\#2.} \textcolor{Bittersweet}{Nếu tình trạng tiến triển thành say nắng, thì da nóng, khô là điển hình  khi các mạch máu giãn ra trong nỗ lực tăng mất nhiệt.} 
\textit{(If the condition progresses to heatstroke, hot and dry skin is typical as blood vessels dilate in an effort to increase heat loss.)} &
\textbf{\#2.} \textcolor{Bittersweet}{Có một mức độ tiêm chủng cao.} 
\textit{(There is a high level of vaccination coverage.)}
& 
\textbf{\#2.} \textcolor{Bittersweet}{Ở trung tâm của Trái Đất, nhiệt độ có thể đạt tới 7000K và áp suất có thể lên tới 360 Gpa.} 
\textit{(At the Earth’s core, temperatures can reach up to 7000 K and pressures can rise to 360 GPa.)} \\ \hline
\textbf{\#3.} \textcolor{Bittersweet}{Caesi có điểm nóng chảy ở 28,4 °C (83,1 °F), là một trong ít các kim loại nguyên tố ở dạng lỏng trong điều kiện gần nhiệt độ phòng.} 
\textit{(Caesium has a melting point of 28.4°C (83.1°F) and is one of the few elemental metals that exist in liquid form under conditions near room temperature.)} &
\textbf{\#3.} \textcolor{Bittersweet}{Nếu tình trạng tiến triển thành say nắng, thì da nóng, khô là điển hình  khi các mạch máu giãn ra trong nỗ lực tăng mất nhiệt.} 
\textit{(If the condition progresses to heatstroke, hot and dry skin is typical as blood vessels dilate in an effort to increase heat loss.)}
& 
\textbf{\#3.} \textcolor{Bittersweet}{Caesi có điểm nóng chảy ở 28,4 °C (83,1 °F), là một trong ít các kim loại nguyên tố ở dạng lỏng trong điều kiện gần nhiệt độ phòng.} 
\textit{(Caesium has a melting point of 28.4°C (83.1°F) and is one of the few elemental metals that exist in liquid form under conditions near room temperature.)} \\
\hline

\end{longtable}
\hl{To complement the quantitative alignment–uniformity results, Table} \ref{tab:Top3-IR} \hl{presents qualitative retrieval examples that illustrate how the learned embedding geometry affects semantic evidence selection. For clarity, we compare ViCLSR with XLM-R$_{Large}$ (the multilingual backbone of our model) and PhoBERT$_{Large}$ (a strong Vietnamese-specific baseline), representing multilingual and monolingual encoder architectures, respectively.} For XLM-R$_{Large}$ and PhoBERT$_{Large}$, the top-ranked sentences often achieve high cosine similarity scores but fail to capture the exact information required to verify the claim. In contrast, ViCLSR consistently retrieves semantically precise evidence, particularly in cases with explicit factual cues such as Query 1 and Query 2. However, more challenging scenarios arise in implicit semantic cases, as illustrated in Query 3 and Query 4, where the model must go beyond surface-level matching to perform deeper reasoning — such as numerical approximation (e.g., “28°C” implying “around 30°C”) or conceptual inference (e.g., “average tidal range” referring to “between high and low tide levels”). These examples highlight the limitations of current models in handling implicit semantics and reasoning-intensive retrieval tasks.

\subsection{Sensitivity to Training Data Size}
\label{training_size}
To further examine the robustness of ViCLSR in low-resource scenarios, we conducted a sensitivity analysis to evaluate how the model performs when the amount of available training data is limited. This experiment aims to assess the stability and scalability of the supervised contrastive learning framework when trained with varying data sizes, which is a common challenge in low-resource language settings.

\begin{figure*}[ht]
    \centering
    \includegraphics[width=1\linewidth]{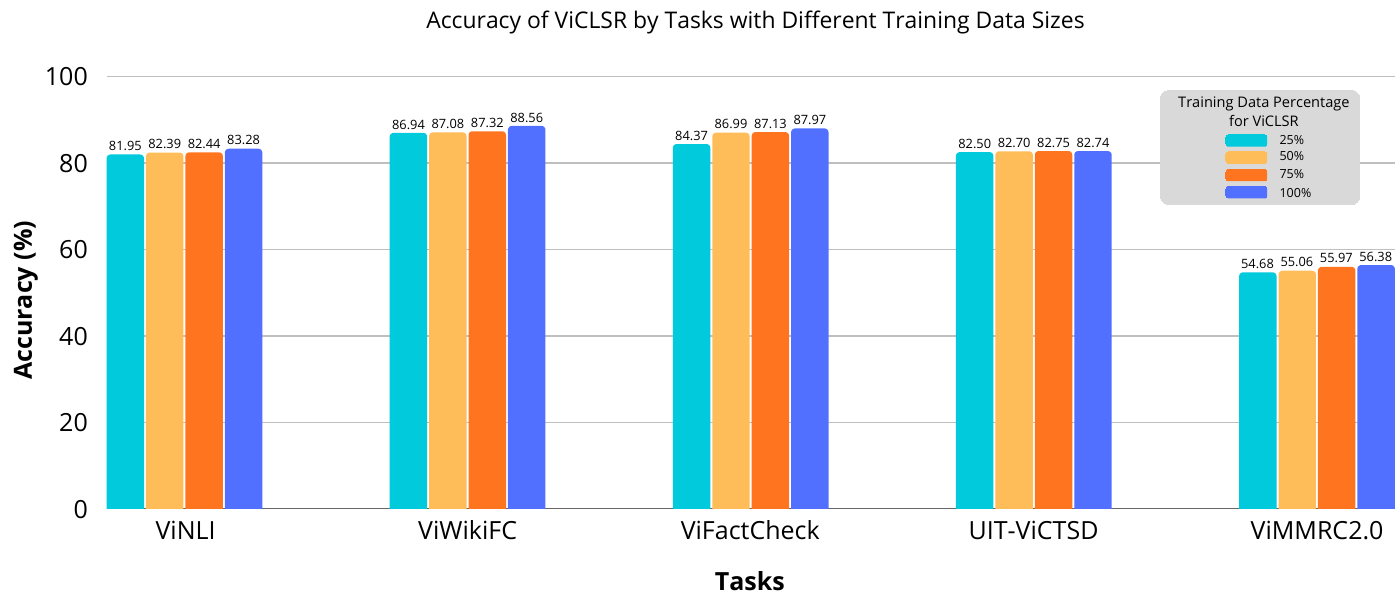}
    \caption{Impact of training data size during contrastive learning on the downstream performance of ViCLSR across five Vietnamese NLU tasks.}
    \label{fig:traing-size}
\end{figure*}

As described in Section \ref{sect:DataPreparation}, the contrastive dataset used for training consists of 8,594 instances derived from Vietnamese NLI corpora. To simulate different levels of data availability, we randomly sampled three subsets corresponding to 25\%, 50\%, and 75\% of the original dataset. The ViCLSR model was trained independently on each subset while keeping all other hyperparameters fixed. After contrastive pretraining, the resulting models were fine-tuned on five downstream NLU tasks, including ViNLI, ViWikiFC, ViFactCheck, UIT-ViCTSD, and ViMMRC2.0. Figure \ref{fig:traing-size} presents the accuracy obtained on the development sets for each task among ViCLSR models trained with different portions of the original dataset and the baseline model trained with the full data.

The results show a consistent upward trend as the amount of training data increases, indicating that ViCLSR effectively leverages additional contrastive supervision to refine semantic representations. More importantly, even with only 25\% of the original data, the model maintains competitive performance across all tasks, suggesting strong generalization capability and robustness to data scarcity. These findings confirm that the proposed framework can function reliably in extreme low-resource conditions and is well-suited for adaptation to other languages with limited annotated resources.

\subsection{Attention Analysis}
\label{attention}

In this analysis experiment, we visualize the attention mechanisms of ViCLSR and the XLM-R model to compare how they handle semantic attention between premise and hypothesis sentences in the NLI task. The goal of this analysis is to evaluate how the models focus on important keywords in the sentences and to highlight the effectiveness of ViCLSR in understanding the semantic relationships between sentences, in comparison to the XLM-R model.

\begin{figure*}[]
  \begin{subfigure}{.5\textwidth}
  \centering
    \includegraphics[width=1\linewidth]{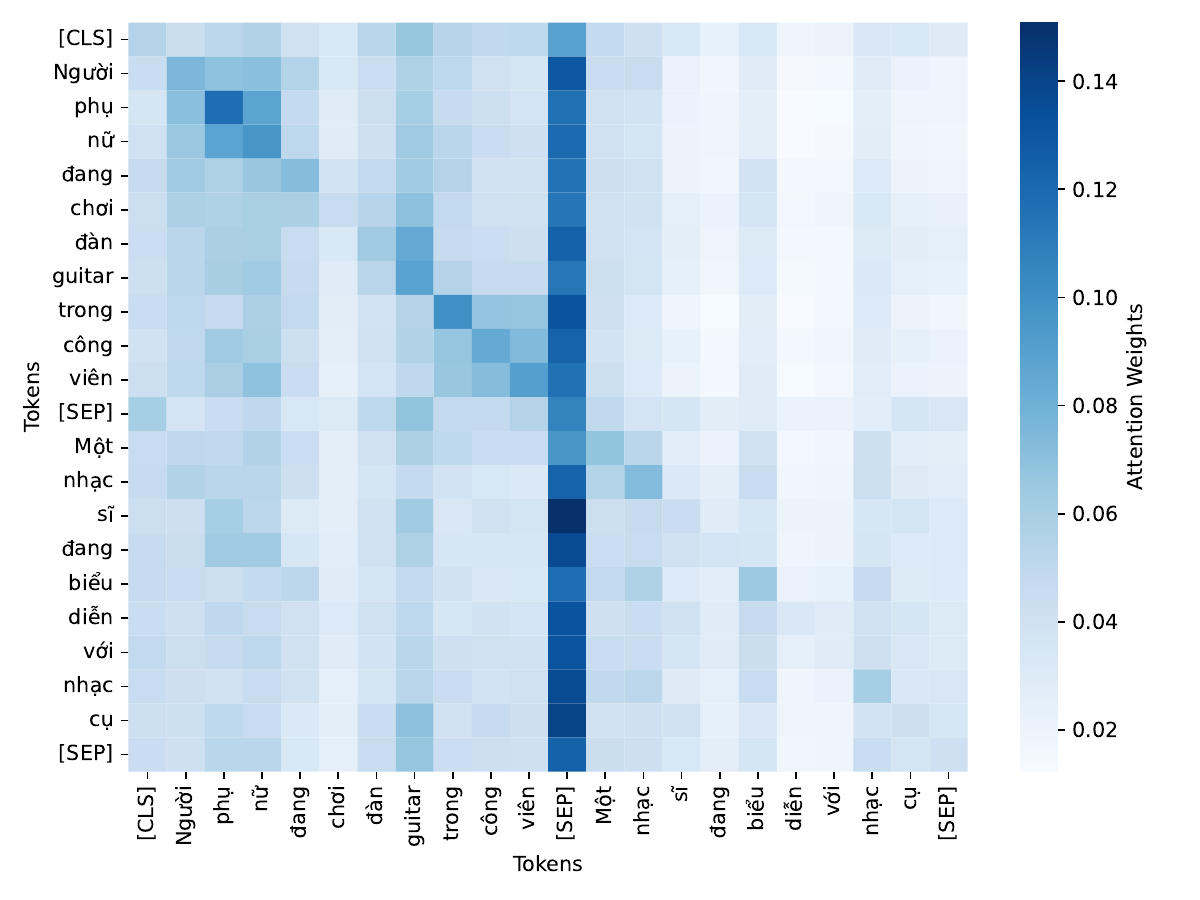}
    \caption{ViCLSR}
    \label{4a}
  \end{subfigure}%
  \begin{subfigure}{.5\textwidth}
  \centering
\includegraphics[width=1\linewidth]{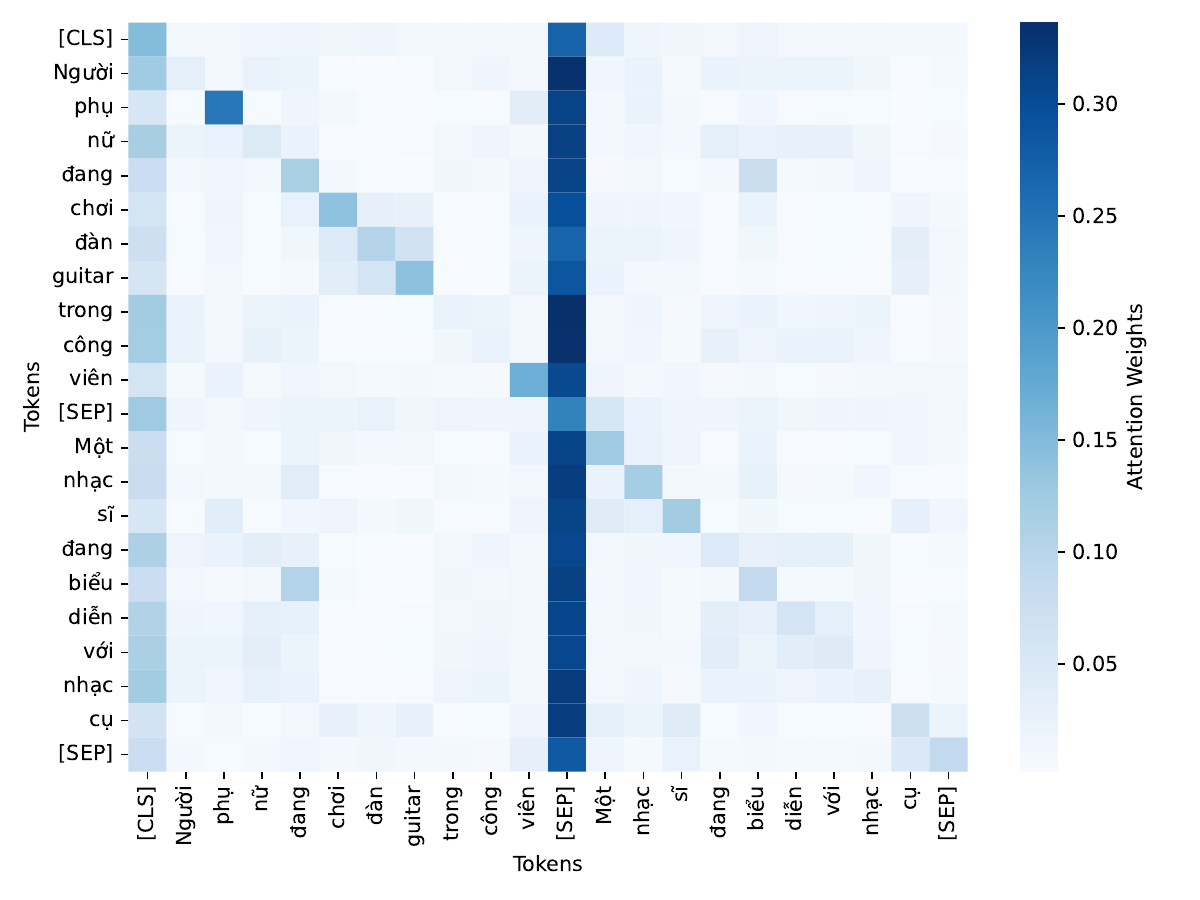}
    \caption{XLM-R}
    \label{4b}
  \end{subfigure}
    \begin{subfigure}{.5\textwidth}
  \centering
    \includegraphics[width=1\linewidth]{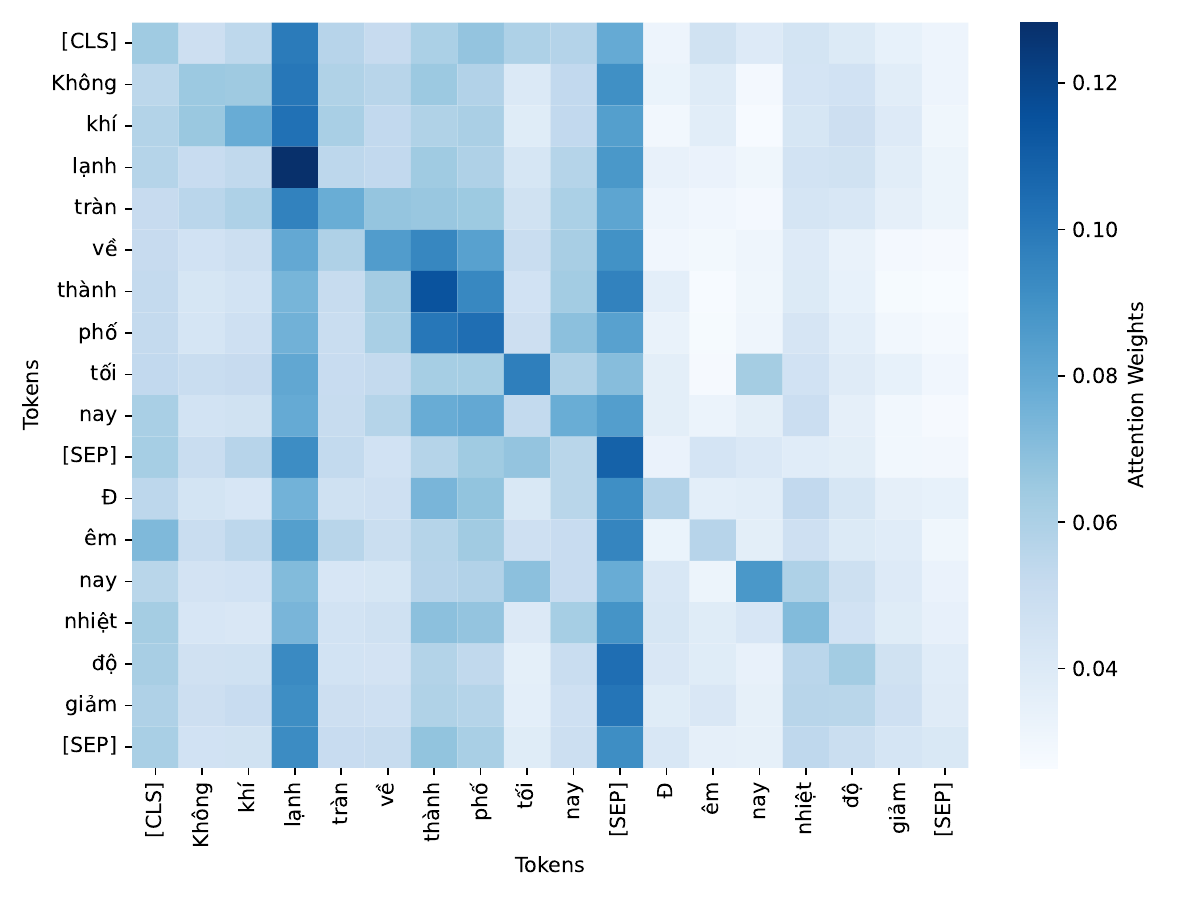}
    \caption{ViCLSR}
    \label{4c}
  \end{subfigure}%
  \begin{subfigure}{.5\textwidth}
  \centering
    \includegraphics[width=1\linewidth]{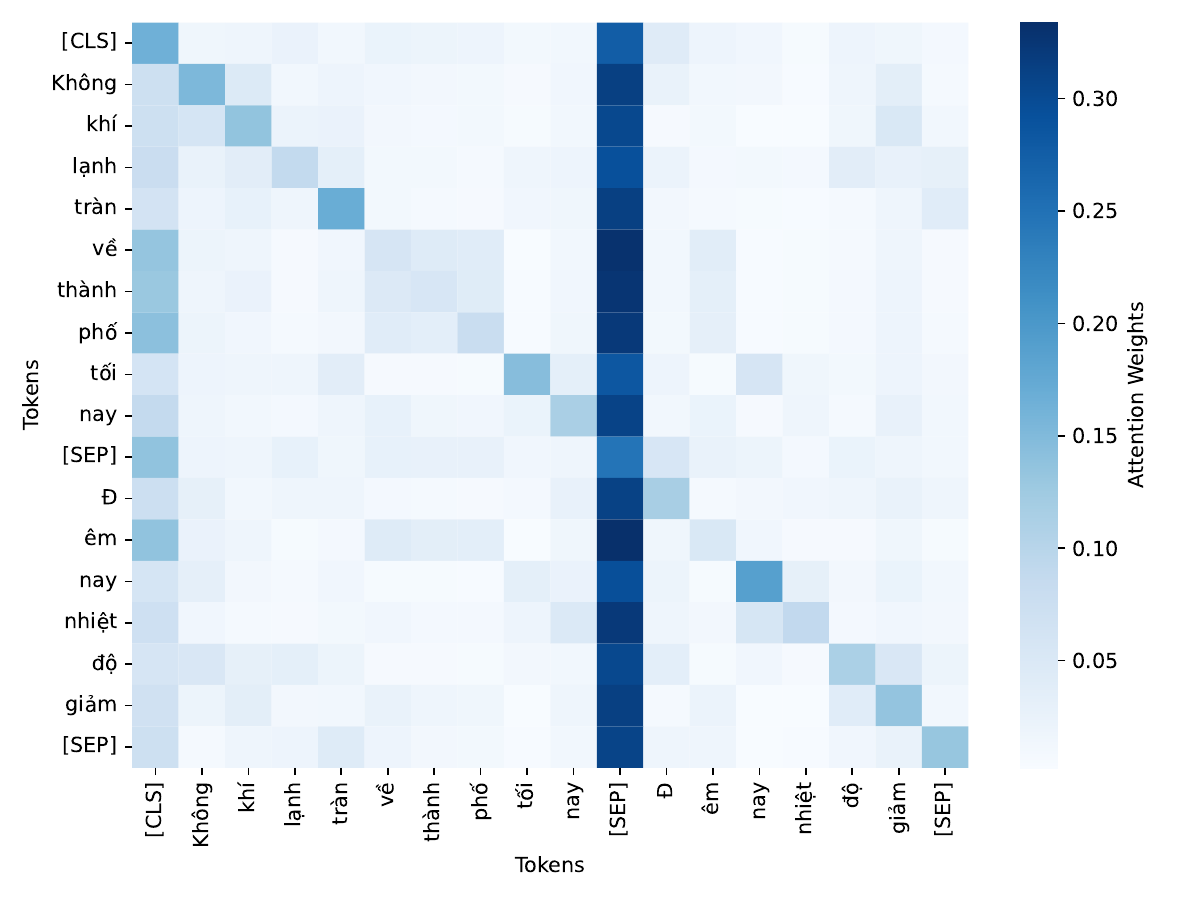}
    \caption{XLM-R}
    \label{4d}
  \end{subfigure}
  \begin{subfigure}{.5\textwidth}
  \centering
    \includegraphics[width=1\linewidth]{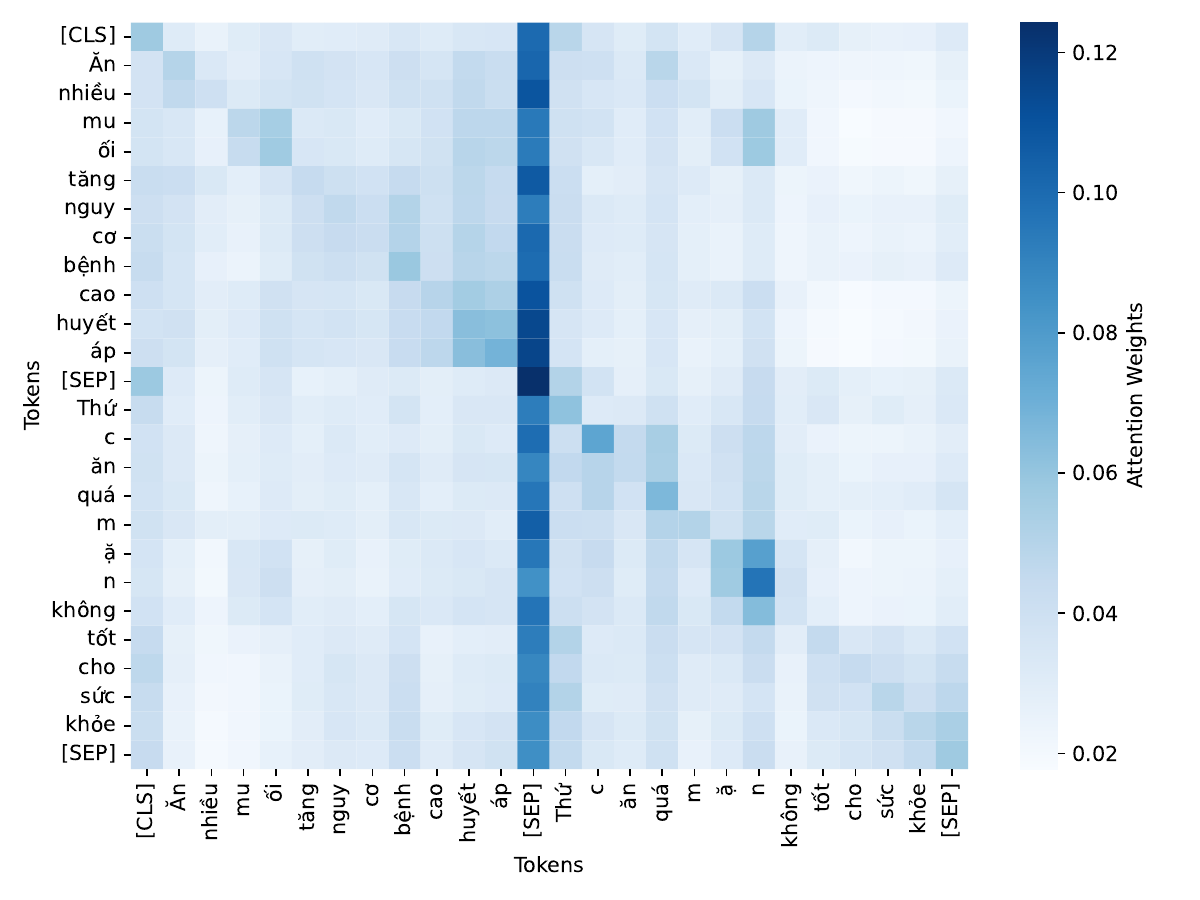}
    \caption{ViCLSR}
    \label{4e}
  \end{subfigure}%
  \begin{subfigure}{.5\textwidth}
  \centering
    \includegraphics[width=1\linewidth]{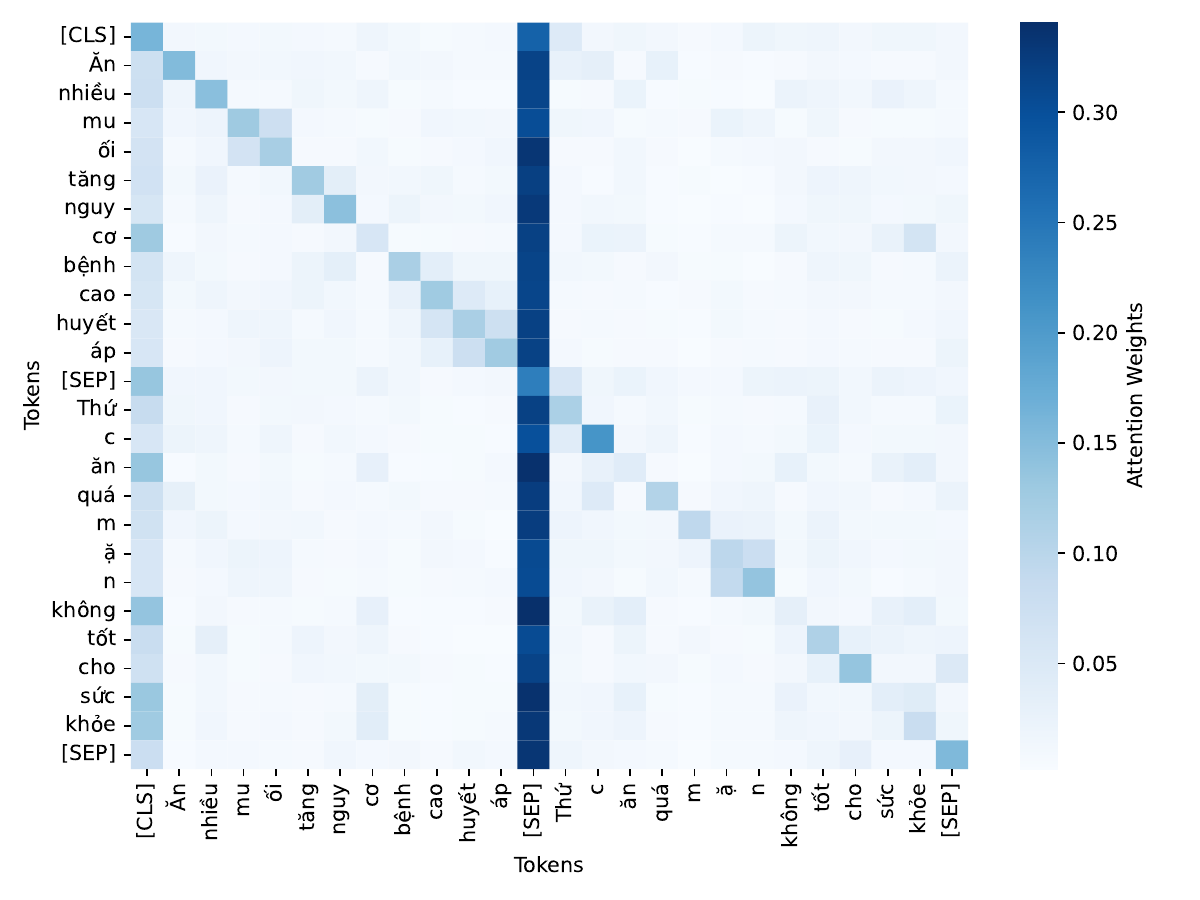}
    \caption{XLM-R}
    \label{4f}
  \end{subfigure}
  \caption{Visualization of the Attention Map Between a Premise and Hypothesis Sentence Using the Pre-Trained XLM-R Model and Our ViCLSR Model.}
  \label{attention_map}
\end{figure*}

The attention visualizations from both models, extracted from the last layer of each as shown in Figure \ref{attention_map}, reveal a clear difference in how they process information. In pairs of sentences such as premise = \textit{“Người phụ nữ đang chơi đàn guitar trong công viên” (A woman is playing guitar in the park)} and hypothesis = \textit{“Một nhạc sĩ đang biểu diễn với nhạc cụ” (A musician is performing with a musical instrument)} in Figures \ref{4a} and \ref{4b}, the ViCLSR model focuses attention on crucial keywords like “\textcolor{violet}{guitar}” and “\textcolor{violet}{nhạc cụ (instrument)}” or “\textcolor{orange}{phụ nữ (woman)}” and “\textcolor{orange}{nhạc sĩ (musician)}”. Additionally, in the pair of sentences premise = \textit{“Không khí lạnh tràn về thành phố tối nay” (Cold air flows into the city tonight)} and hypothesis = \textit{“Đêm nay nhiệt độ giảm” (Tonight the temperature drops)}, as seen in Figures \ref{4c} and \ref{4d}, the attention of the ViCLSR model is predominantly on the words “\textcolor{olive}{lạnh (cold)}” and “\textcolor{olive}{nhiệt độ giảm (temperature drops)}”. In the pair premise = \textit{“Ăn nhiều muối tăng nguy cơ bệnh cao huyết áp” (Eating a lot of salt increases the risk of high blood pressure disease)} and hypothesis = \textit{“Thức ăn quá mặn không tốt cho sức khỏe” (Too salty food is not good for health)}, as shown in Figures \ref{4e} and \ref{4f}, the words “\textcolor{teal}{muối (salt)}” and “\textcolor{teal}{mặn (salty)}” as well as “\textcolor{purple}{bệnh (disease)}” and “\textcolor{purple}{không tốt cho sức khỏe (not good for health)}” receive more attention from the ViCLSR model. In contrast, the attention from the XLM-R model is less clearly demonstrated in the visualizations, indicating that ViCLSR has a stronger ability to understand semantics compared to XLM-R.

ViCLSR places more emphasis on words with strong semantic connections between the premise and hypothesis, enabling the model to more accurately identify the similarities and differences between the sentences. In contrast, the XLM-R model distributes attention more broadly and does not focus on specific words, instead highlighting the overall context of the sentence. This suggests that while XLM-R can understand the global meaning of a sentence, it lacks the nuance required to attend to specific semantic relationships between the sentences. Therefore, ViCLSR proves superior in tasks requiring semantic relationship understanding, thanks to its ability to learn and recognize clearer semantic connections.

\subsection{Recommendation for Other Low Resource Languages}
\label{recommend}

For low-resource languages, the key challenge lies in the availability of annotated datasets, which are crucial for training supervised contrastive learning models. In the case of Vietnamese, we utilized existing NLI datasets to create positive and negative sentence pairs, which are essential for learning effective sentence embeddings. The ViCLSR framework can be applied to other low-resource languages in the same way: by identifying or creating NLI-style datasets that include sentence pairs labeled with entailment and contradiction. This flexibility of ViCLSR allows it to handle a variety of low-resource languages, such as Thai, Indonesian, Burmese, and Lao, as long as the key requirement that positive and negative sentence pairs are met. If NLI datasets are not available, alternative datasets such as paraphrase or fact checking datasets can be repurposed to fit this task. Additionally, augmenting the dataset with synthetic data generated through machine translation, sentence paraphrasing, or other augmentation techniques ensures a more diverse and rich training set, further enhancing performance in low-resource settings.

In addition to dataset preparation, the choice of pre-trained multilingual models plays a critical role in transferring knowledge from high-resource languages to low-resource ones. By leveraging models like XLM-R, which have been pre-trained on a vast multilingual corpus, the ViCLSR framework can be fine-tuned on the NLI dataset of the target language. This fine-tuning process significantly improves the ability of the model to understand the unique semantic relationships of the target language, even when there is limited labeled data. The adaptability of ViCLSR makes it a powerful tool for low-resource languages, as it capitalizes on the strength of multilingual pre-trained models and applies contrastive learning to improve sentence embeddings. For languages with particularly complex syntax or semantic nuances, adjusting the model architecture or the data preparation process is necessary. Ultimately, the success of applying the ViCLSR framework to other low-resource languages depends on the careful preparation of data, the use of suitable pre-trained models like XLM-R, and ongoing research to refine the framework for specific linguistic characteristics.

\section{Conclusion and Future Works}
\label{sect:conclusion}


In this article, we introduced ViCLSR, a supervised contrastive learning model specifically designed to optimize Vietnamese sentence embeddings. The core contribution of this work lies in adapting existing Vietnamese NLI datasets for contrastive learning, enabling the creation of high-quality sentence embeddings tailored to the linguistic characteristics of Vietnamese. Our experiments demonstrated that ViCLSR consistently outperforms both monolingual and multilingual pre-trained models, including PhoBERT, mBERT, XLM-R, and CafeBERT, as well as a contrastive learning baseline DiffCSE, across various Vietnamese NLU tasks such as natural language inference, fact checking, constructive speech detection, and machine reading comprehension.

In particular, ViCLSR surpassed the strong multilingual XLM-R$_{Large}$ model by 1.5\% accuracy on ViNLI, exceeded the monolingual PhoBERT$_{Large}$ by over 6.9\% accuracy, and outperformed the contrastive DiffCSE model by more than 24\% accuracy on the same benchmark. These substantial improvements highlight the effectiveness of our supervised contrastive framework in capturing fine-grained semantic relationships and complex reasoning patterns. Furthermore, the analysis of sentence embedding distributions reveals that ViCLSR produces embeddings that are both well-aligned and uniformly distributed, demonstrating the robustness of contrastive learning in generating high-quality and semantically coherent representations for Vietnamese.

In future, further adaptation of contrastive learning techniques will be explored to fully leverage the unique linguistic features of Vietnamese. In particular, extending ViCLSR with advanced mechanisms such as adaptive negative sampling and dynamic temperature adjustment represents a promising direction that could further enhance performance. \hl{Additionally, a primary focus of future research will be transitioning from encoder-only architectures to sequence-to-sequence frameworks to address generative tasks such as abstractive summarization, machine translation, and dialogue systems. Our supervised contrastive learning strategy will be integrated with Vietnamese-specific generative models like BARTpho} \cite{bartpho} \hl{and ViT5} \cite{phan2022vit5}. \hl{ViCLSR’s high-quality embeddings can enhance Retrieval-Augmented Generation (RAG) for Vietnamese-centric LLMs such as GPT} \cite{achiam2023gpt} \hl{and DeepSeek} \cite{liu2024deepseek}. Finally, image captioning datasets (e.g., UIT-ViIC \cite{lam2020uit}) can be leveraged  to explore multimodal contrastive learning enhanced with adversarial data from ViANLI \cite{van2024vianli}, aligning Vietnamese text with visual semantics.

\section*{Acknowledgement}
This research is funded by Vietnam National University HoChiMinh City (VNU-HCM) under grant number DS.C2025-26-10.

\section*{Declarations}

\textbf{Conflict of interest} The authors declare that they have no conflict of interest.

\section*{Data Availability}

No datasets were generated or analyzed during the current study.


\section*{Author Contribution}
Tin Van Huynh: Conceptualization; Formal analysis; Investigation; Methodology; Validation; Visualization; Writing - original draft.

Kiet Van Nguyen: Conceptualization; Formal analysis; Investigation; Validation; Visualization; Supervision; Writing - review\&editing.

Ngan Luu-Thuy Nguyen: Conceptualization; Formal analysis; Investigation; Methodology; Validation; Supervision; Writing - review\&editing.

\bibliography{sn-bibliography}

\appendix

\section{Analysis of Model Stability and Convergence} 
\label{loss_curve}

\begin{figure*}[]
  \begin{subfigure}{.5\textwidth}
  \centering
    \includegraphics[width=1\linewidth]{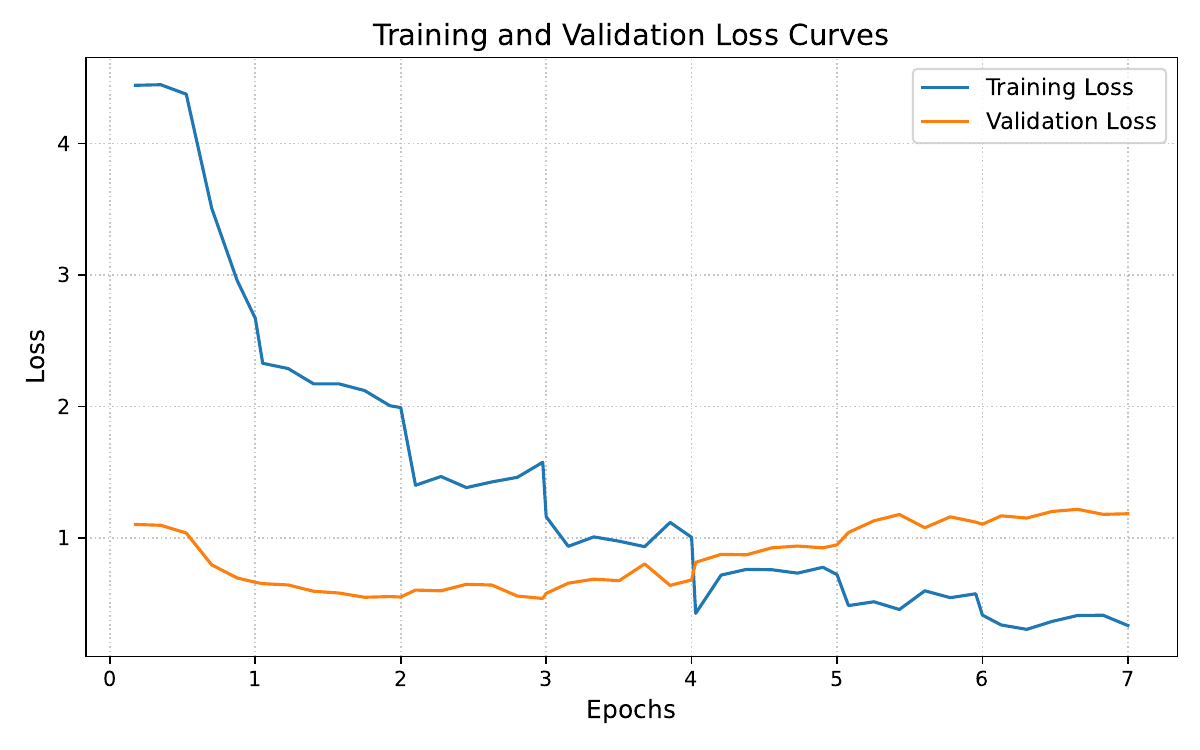}
    \caption{ViNLI}
    \label{loss_vinli}
  \end{subfigure}%
  \begin{subfigure}{.5\textwidth}
  \centering
\includegraphics[width=1\linewidth]{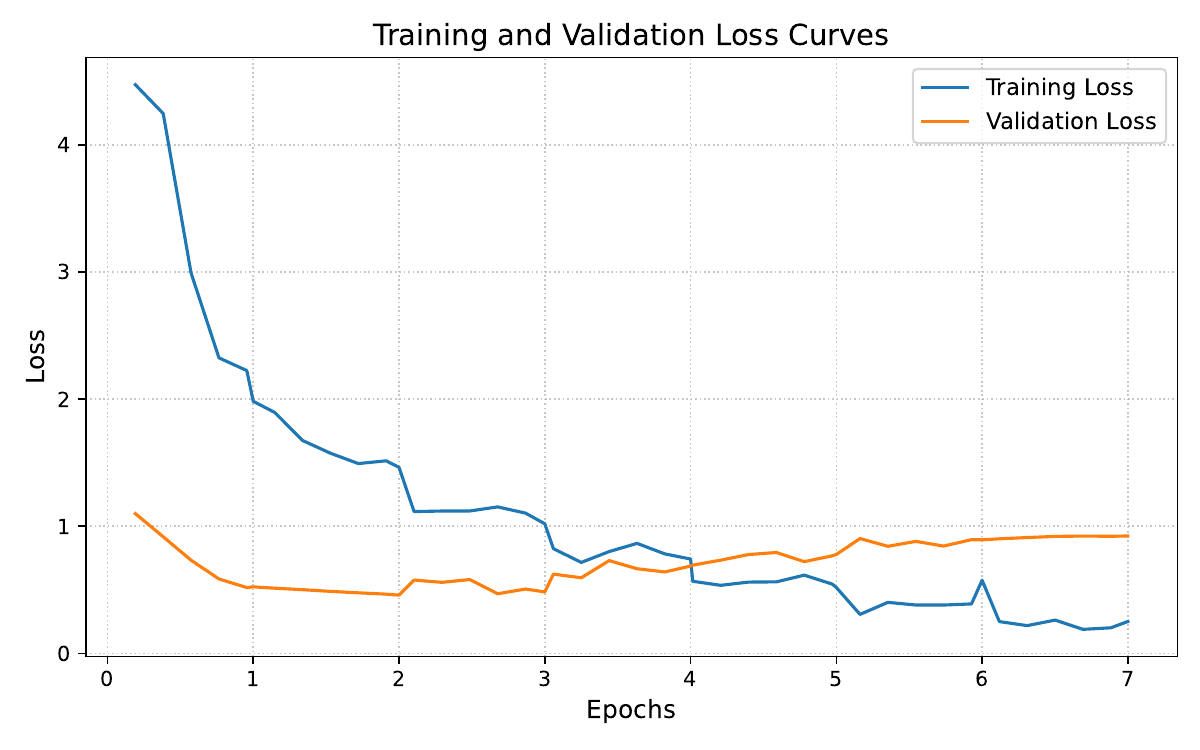}
    \caption{ViWikiFC}
    \label{loss_viwikifc}
  \end{subfigure}
    \begin{subfigure}{.5\textwidth}
  \centering
    \includegraphics[width=1\linewidth]{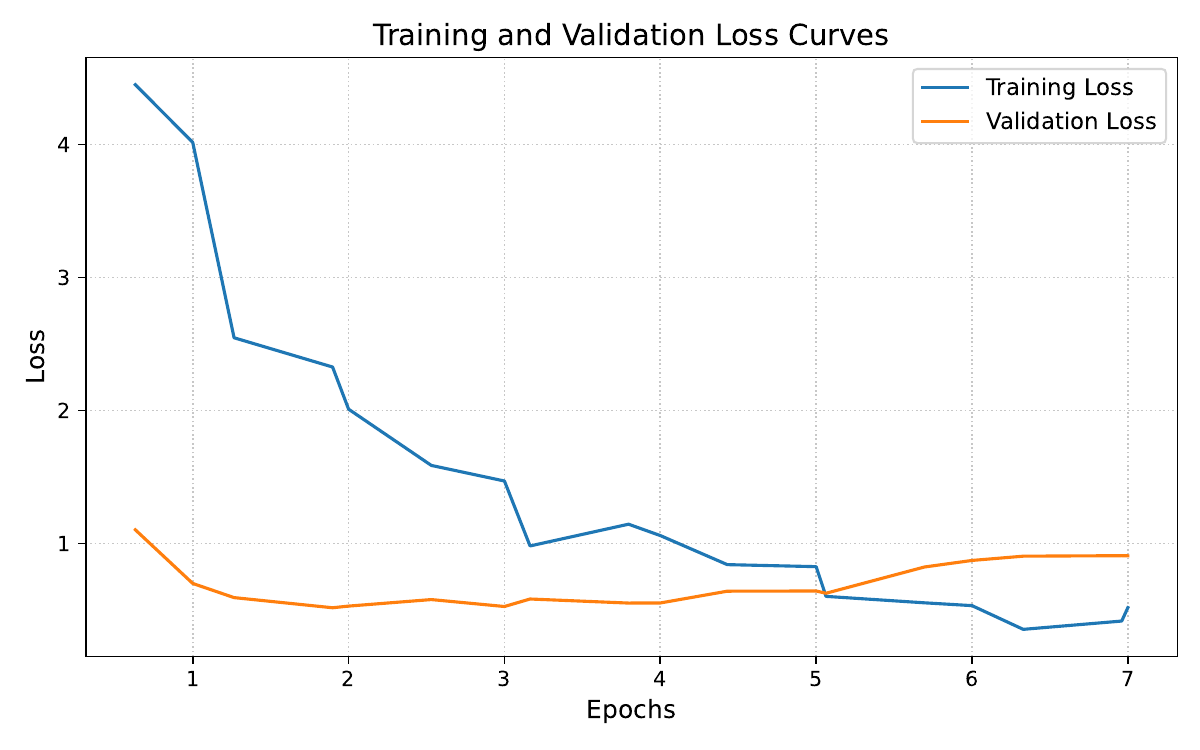}
    \caption{ViFactCheck}
    \label{loss_vifactcheck}
  \end{subfigure}%
  \begin{subfigure}{.5\textwidth}
  \centering
    \includegraphics[width=1\linewidth]{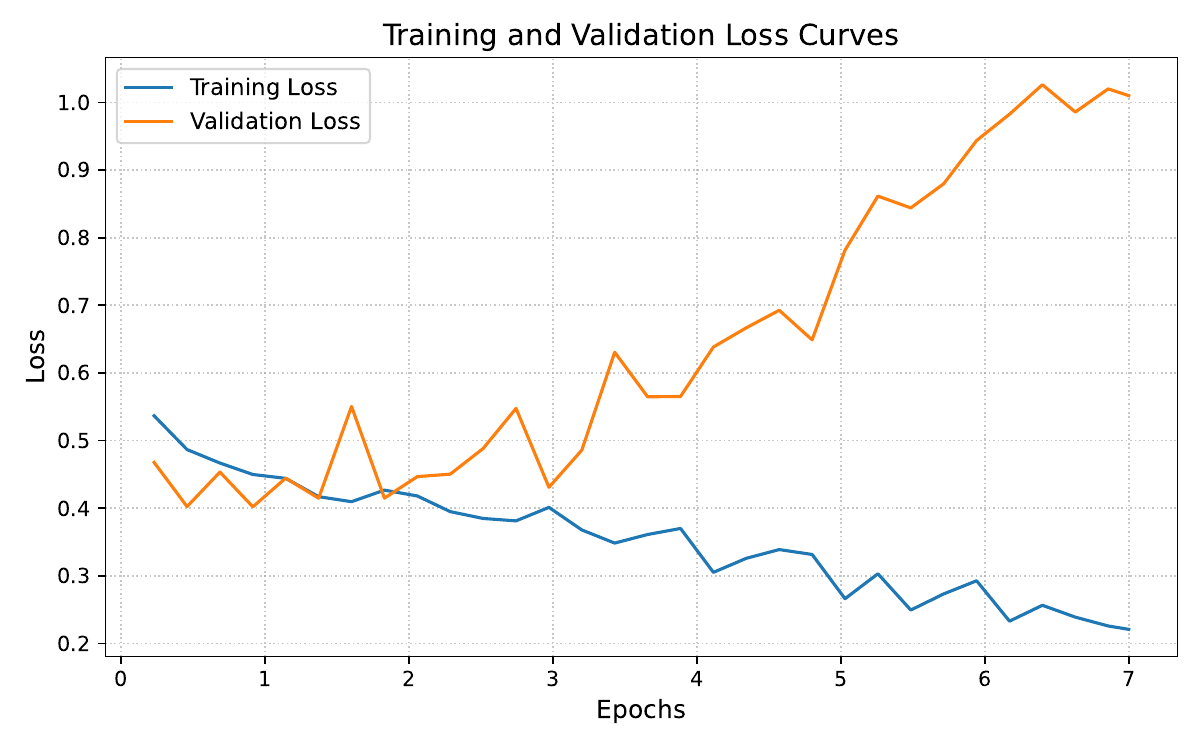}
    \caption{UIT-ViCTSD }
    \label{loss_uit_victsd}
  \end{subfigure}
  \centering
  \begin{subfigure}{.5\textwidth}
  \centering
    \includegraphics[width=1\linewidth]{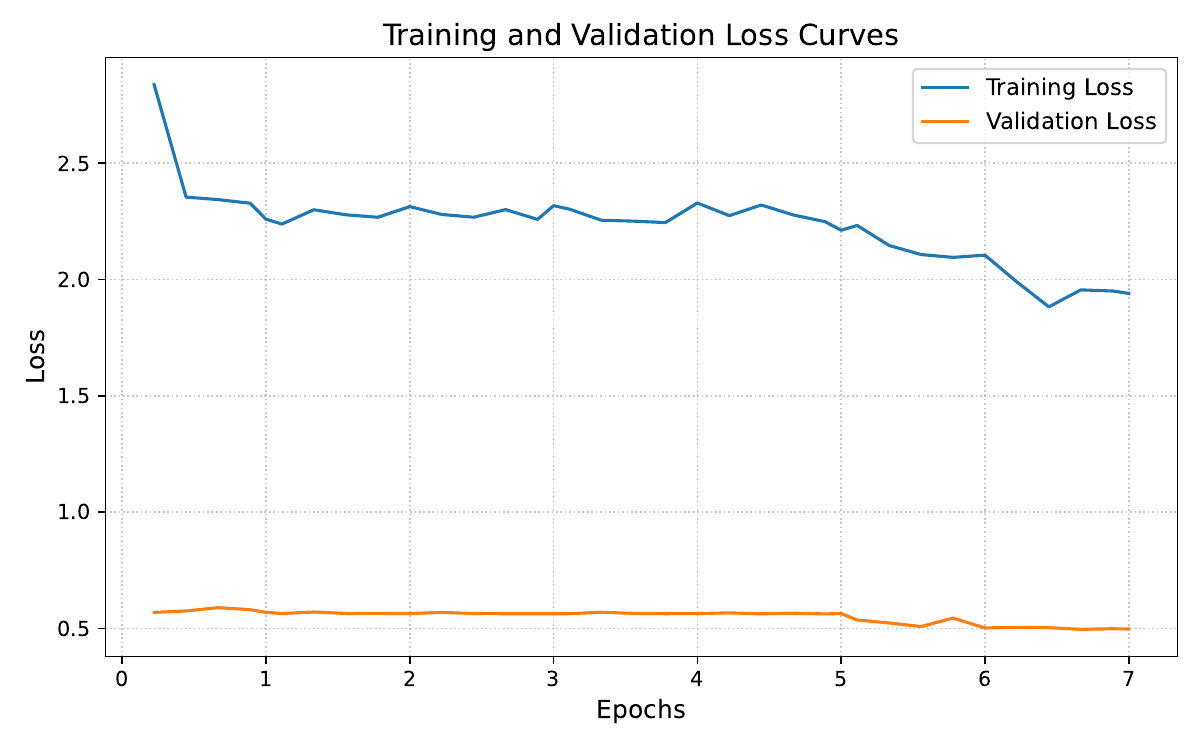}
    \caption{ViMMRC2.0}
    \label{loss_vimmrc2.0}
  \end{subfigure}%
  \caption{\hl{Training-validation loss curves of the ViCLSR model on downstream tasks.}}
  \label{loss_curve_fig}
\end{figure*}

\hl{To explicitly address potential overfitting risks and provide empirical support for our experimental rigor, Figure} \ref{loss_curve_fig}, \hl{presents the training and validation loss curves across all evaluated tasks. While the training loss steadily decreases across all datasets, the validation trajectories reveal distinct convergence patterns based on task complexity and data scale.

For natural language inference and fact-checking (ViNLI, ViWikiFC, and ViFactCheck; Figure} \ref{loss_vinli}, \ref{loss_viwikifc}, and \ref{loss_vifactcheck}), \hl{the validation loss exhibits a standard and highly stable convergence trend. It reaches an optimal minimum within the early epochs before plateauing or showing a very slight upward drift. This indicates a robust learning process where the model captures semantic relationships and generalizes well to unseen data before any significant memorization occurs.

For constructive speech detection (UIT-ViCTSD; Figure} \ref{loss_uit_victsd}), \hl{the training loss demonstrates a smooth and consistent downward trend throughout the training process, indicating effective optimization of the supervised contrastive objective. In contrast, the validation loss initially remains stable but begins to increase gradually in later epochs. This divergence suggests a tendency toward overfitting when training continues excessively. However, the best-performing model checkpoint is selected based on validation performance prior to this divergence, ensuring that the final reported results correspond to the most generalizable model state. This behavior reflects a common pattern in moderately sized classification datasets, where semantic boundary learning stabilizes early while prolonged training may lead to marginal memorization effects.

For the machine reading comprehension task (ViMMRC2.0; Figure} \ref{loss_vimmrc2.0}), \hl{which has a relatively smaller training size and higher reasoning complexity, the validation loss remains remarkably stable across epochs with only minor fluctuations. Notably, no sharp divergence between training and validation curves is observed. Although the training loss continues to decrease steadily, the validation loss maintains a nearly flat trajectory, indicating that the model does not excessively overfit to the training distribution. This empirical evidence contradicts the concern of severe overfitting on small datasets and demonstrates that the supervised contrastive framework maintains stable generalization behavior even under limited data conditions.}

\end{document}